\documentclass[10pt,twocolumn,letterpaper]{article}

\usepackage[pagenumbers]{cvpr}

\usepackage{microtype}

\usepackage{graphicx}
\usepackage{booktabs}
\usepackage{amsmath}
\usepackage{amssymb}
\usepackage{algorithm}
\usepackage{algpseudocode}

\newcommand{\neff}{n_{\mathrm{eff}}}

\newcommand{\kstar}{k^{*}}
\newcommand{\etat}{\eta_t}
\newcommand{\tauc}{\tau_c}
\newcommand{\Wsum}{W_{\Sigma}}
\newcommand{\Wdiff}{W_{\Delta}}

\renewcommand{\topfraction}{0.9}
\renewcommand{\dbltopfraction}{0.9}
\renewcommand{\textfraction}{0.1}

\usepackage{colortbl}
\definecolor{cvprblue}{rgb}{0.21,0.49,0.74}
\usepackage[pagebackref,breaklinks,colorlinks,allcolors=cvprblue]{hyperref}
\hypersetup{pdftitle={Tokenize Moments, Not Frames: Tokenizer-Aware Budget Allocation for Training-Free Video LLM Acceleration},pdfauthor={Zirui Zhao, Kai Liu, Linghe Kong}}
\makeatletter
\newcommand{\appendixtoc}{%
  \par\noindent{\bfseries Table of Contents}\par\nobreak
  \vspace{2pt}\hrule\vspace{1pt}%
  {\small\hypersetup{linkcolor=black}%
   \let\cvprnumberline\numberline
   \def\numberline##1\hskip-1em.##2{\cvprnumberline{##1}\ifx##2\nobreakspace\expandafter\@gobble\fi}%
   \@starttoc{toc}\par}%
  \vspace{6pt}\hrule\vspace{10pt}}
\makeatother

\title{Tokenize Moments, Not Frames:\\ Tokenizer-Aware Budget Allocation for Training-Free Video LLM Acceleration}

\author{Zirui Zhao$^{1}$\qquad Kai Liu$^{2}$\qquad Linghe Kong$^{2}$\thanks{Corresponding author: Linghe Kong, linghe.kong@sjtu.edu.cn}\\
$^{1}$Huazhong University of Science and Technology\qquad
$^{2}$Shanghai Jiao Tong University}

\begin{document}
\addtocontents{toc}{\protect\setcounter{tocdepth}{0}}
\setcounter{footnote}{1}
\maketitle
\begin{abstract}
Video large language models (LLMs) represent each video with thousands of visual tokens, and training-free accelerators that prune or merge these tokens lose accuracy on long videos.
Analyzing frame similarity and the vision tokenizers of 16 open models, we trace this loss to a step before token selection.
Under a token budget, the frames of a long video are sampled farther apart than the roughly 2\,s over which frames remain similar.
The two-frame 3D patches of models such as Qwen-VL then blend two unrelated scenes into each visual token, a \emph{double exposure} that arises before any pruning or merging.
Fixed-size tokenizers such as InternVL's avoid this blend but respond differently to the same reduction.
We propose ESSENCE, a training-free method that tokenizes \emph{moments}, segments within which frames remain similar, rather than frames.
Before encoding, a closed-form rule uses the sampling interval and the tokenizer class to set the number of token grids and the dimension to reduce; visual tokens are then released from later LLM layers.
Against five recent accelerators on Qwen3-VL, Qwen2.5-VL and InternVL3.5, ESSENCE achieves higher accuracy in 58 of 60 comparisons at matched mean time to first token (TTFT).
With 80\% of the tokens, it matches the average accuracy of the full Qwen models at $1.6\times$ faster TTFT; on GLM-4.1V, it exceeds the full model by up to 3.4 points.
Code is available at: \url{https://github.com/zhkandsrz/ESSENCE}.
\end{abstract}

\section{Introduction}
\label{sec:intro}

Video large language models (LLMs) have advanced complex video understanding~\cite{bai2025qwen3,bai2025qwen25,zhu2025internvl3,glm2025v}, but their computational cost limits deployment.
On Qwen3-VL, 32 frames enter the LLM as about 3000 visual tokens, which account for most of the time to first token (TTFT).
A deployment can therefore afford only a limited token budget, which bounds the number of frames regardless of the video length.

Inside the LLM, FastV~\cite{chen2024image} prunes low-attention tokens, ShortV~\cite{yuan2025shortv} freezes visual tokens in low-contribution layers, and VTW~\cite{lin2025boosting} withdraws them in late layers.
After the vision encoder, DyCoke~\cite{zhao2025dycoke}, FastVID~\cite{shao2025fastvid}, HoliTom~\cite{shao2025holitom} and FlashVID~\cite{cao2026flashvid} merge or prune similar tokens across frames.
iLLaVA~\cite{hu2026illava} and EarlyTom~\cite{earlytom2026} move this compression into the encoder, and LoHi~\cite{dong2026lohi} lowers the input resolution before encoding to afford more frames.
Two problems remain: accuracy degrades on long videos, and post-encoder methods leave the encoding latency unchanged.
On videos sampled more than 2\,s apart, all five methods we measured (FastVID, FlashVID, iLLaVA, EarlyTom and ShortV) fall significantly below the full model, by 1.4 to 6.4 percentage points.

\begin{figure*}[t]
\centering
\includegraphics[width=0.95\textwidth]{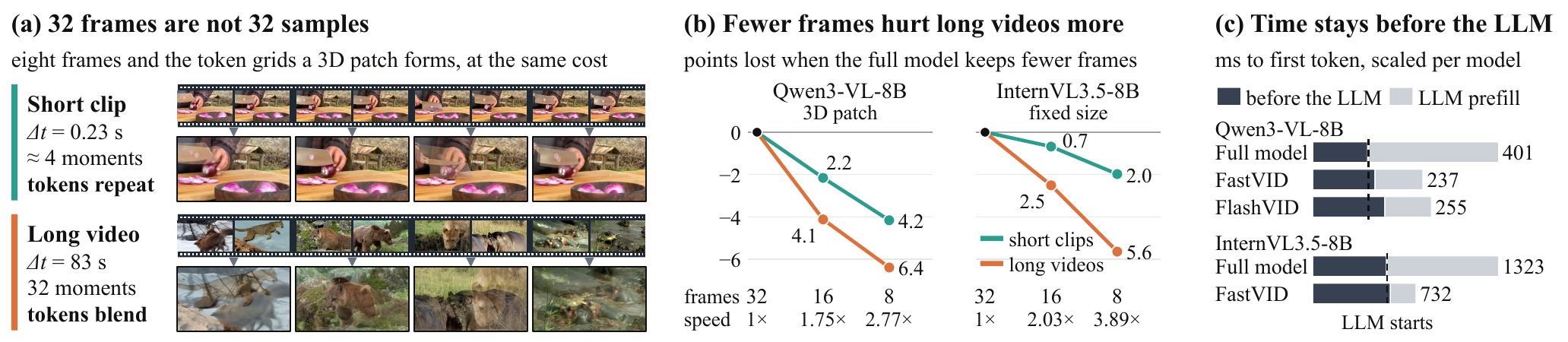}
\caption{32 frames are not 32 samples. (a) Eight consecutive frames and the token grids of a two-frame patch (pixel mean shown) on Qwen3-VL-8B: redundant on a short clip, blended on a long video, at about the same token cost. (b) Accuracy loss when the full model retains 16 or 8 of its 32 frames, on short clips ($\Delta t\le1$\,s) and long videos ($\Delta t>2$\,s), with the TTFT speedup; on Qwen-VL, fewer frames also means fewer grids. (c) TTFT before and inside the LLM, normalized per model (labels in ms; dashed: LLM start in the full model).}
\label{fig:intro}
\end{figure*}

To understand why, we examine what the visual tokens of a long video contain.
Frames more than about 2\,s apart are nearly as dissimilar as frames from different videos (\cref{sec:moments}); we call a segment whose frames remain correlated a \emph{moment}, and a video \emph{sparse} when its frames are farther apart than this.
Since the budget bounds the frame count, long videos are sparse: even 256 frames are 14\,s apart on a one-hour video.
At the same token cost, the median short clip contains five to six moments and the median long video 32 (\cref{fig:moments}b), and removing half of the frames costs Qwen3-VL and InternVL two to four times more accuracy on long videos than on short clips (\cref{fig:intro}b).

Inspecting the patch kernels and preprocessing configurations of 16 open models, we find two classes of vision tokenizers.
Eight models, such as Qwen-VL and GLM-4.1V, encode every two frames into one token grid with a 3D patch embedding~\cite{arnab2021vivit} whose two temporal kernels are nearly identical, which effectively sums the two frames.
On a sparse video, this sum blends two unrelated scenes into one grid, like a double exposure on film; encoding each frame independently, at twice the token count, improves accuracy on such videos by 2.4 to 3.8 points on three of these models.
The other eight, such as InternVL and LLaVA-OneVision, encode each frame independently but at a single input resolution.
The least costly reduction differs by class: on densely sampled videos at the same budget, lowering the resolution costs Qwen-VL no more than removing frames, whereas InternVL should retain its resolution and remove frames.

Profiling each stage further shows that although FastVID shortens the LLM prefill by 64\% on Qwen3-VL, 57\% of the remaining TTFT is spent before the LLM, and 76\% on InternVL (\cref{fig:intro}c).

Based on these observations, we introduce ESSENCE, a training-free method that tokenizes moments rather than frames.
It estimates the number of moments from the sampling interval and allocates the token budget before tokenization with a closed-form rule specific to the tokenizer class.
For sparse videos, the rule retains all frames and encodes each one independently; for densely sampled videos, it lowers the spatial resolution on 3D-patch models and the number of frames on fixed-size models (\cref{tab:opposite}).
Where it reduces the encoder input, the allocation also shortens the stage before the LLM.
ESSENCE then \emph{releases} visual tokens from the later LLM layers, where they rarely affect the prediction, and no longer processes them there.

We evaluate ESSENCE under a unified timing protocol on Qwen3-VL, Qwen2.5-VL and InternVL across four benchmarks (TempCompass~\cite{liu2024tempcompass}, MVBench~\cite{li2024mvbench}, Video-MME~\cite{fu2024videomme} and LongVideoBench~\cite{wu2024longvideobench}).
On sparse videos, uniform downscaling to the same budget, the low-resolution stream of LoHi, falls significantly below the full model, whereas ESSENCE shows no significant loss at about $2\times$ faster TTFT.
It also achieves higher accuracy than the five methods above in 58 of 60 comparisons at matched mean TTFT (46 significantly), and on GLM-4.1V it exceeds the full model by up to 3.4 points with $2.2\times$ faster TTFT.

Our contributions are:
\begin{itemize}
  \item We show that under a token budget, the frames of a long video are no longer redundant, and that 3D-patch and fixed-size tokenizers call for different reductions.
  \item We propose ESSENCE, a training-free, closed-form method that allocates the token budget before tokenization by sampling interval and tokenizer class, and releases visual tokens from later LLM layers.
  \item We validate ESSENCE on three models and four benchmarks against five re-measured baselines, and extend it to GLM-4.1V and LLaVA-OneVision.
\end{itemize}

\section{Related Work}
\label{sec:related}

\noindent\textbf{Token compression.}
ToMe~\cite{bolya2023tome} merges similar tokens in vision transformers, and TempMe~\cite{shen2025tempme} across video clips.
For image LLMs, LLaVA-PruMerge~\cite{shang2025prumerge} and VisionZip~\cite{sun2025visionzip} keep high-attention tokens and merge the rest.
For video LLMs, DyCoke~\cite{zhao2025dycoke} merges similar tokens across frames, STTM~\cite{hyun2025multi} merges multi-granular spatiotemporal tokens, and PruneVid~\cite{huang2024prunevid} further prunes question-irrelevant tokens.
FastVID~\cite{shao2025fastvid} and HoliTom~\cite{shao2025holitom} compress within temporal segments, FlashVID~\cite{cao2026flashvid} merges along spatiotemporal trees, and UniComp~\cite{unicomp2025} and AdaReTaKe~\cite{wang2025adaretake} allocate budgets by token uniqueness or redundancy.
iLLaVA~\cite{hu2026illava} and EarlyTom~\cite{earlytom2026} move compression into the vision encoder.
However, all these methods act at or after the patch embedding, where a 3D patch has already summed two frames.

\noindent\textbf{Computation inside the LLM.}
FastV~\cite{chen2024image} and PyramidDrop~\cite{xing2025pyramiddrop} drop visual tokens at early or progressive layers, ShortV~\cite{yuan2025shortv} freezes them in low-contribution layers, and Op-Skip~\cite{opskip2026} and DOP~\cite{dop2025} prune individual operations.
VTW~\cite{lin2025boosting} withdraws visual tokens in late layers, in line with evidence on images that visual information reaches question tokens in early and middle layers~\cite{zhang2025cross}.
Our release step follows this line, with a single relative depth across models and a label-free bound on the accuracy change.

\noindent\textbf{Frame and resolution allocation.}
Closely related methods act before encoding.
AKS~\cite{tang2025adaptive}, Q-Frame~\cite{zhang2025qframe}, LDDR~\cite{lddr2026} and DAFS~\cite{wang2026dafs} select frames, and sometimes resolutions, based on the question, whereas ResAdapt~\cite{resadapt2026} and VideoRouter~\cite{videorouter2026} train modules to allocate the budget.
Concurrent work LoHi~\cite{dong2026lohi} pairs a dense low-resolution stream with sparse high-resolution frames, and a recent study~\cite{khatri2026select} reinvests tokens saved by compression into additional frames.
Our allocation is instead derived in closed form from the sampling interval and the tokenizer, without a selector or training; to our knowledge, no prior work accounts for 3D patches or for reduction costs that depend on the tokenizer.

\section{What Visual Tokens Contain}
\label{sec:mismatch}

We analyze how quickly video frames decorrelate and how vision tokenizers process them, and derive three observations that guide ESSENCE.

\subsection{Preliminaries}
\label{sec:prelim}

A video LLM answers a question from $T$ frames $F=\{f_i\}_{i=1}^{T}$ sampled uniformly at median interval $\Delta t$.
A vision tokenizer maps these frames to visual tokens:
\begin{equation}
  Z = \Pi\big(E(U(F))\big),
  \label{eq:tokenizer}
\end{equation}
where $U$ is the patch embedding, $E$ the vision encoder and $\Pi$ the projector.
The patch embedding groups $u$ consecutive frames at a time, each unit producing one \emph{token grid} of $b$ visual tokens.
With $k$ grids, the LLM receives $kb$ visual tokens; natively, $k=T/u$.

An $L$-layer LLM then processes all tokens; its prefill (the first forward pass over the prompt) accounts for 70\% of TTFT on Qwen3-VL.
Given $T$, $\Delta t$ and a budget $kb\le B$, we aim to choose $k$, $b$, the frames in each grid, and the layer at which visual tokens are removed to maximize accuracy at a target TTFT without training or labels.

\subsection{Frames Are Not Moments}
\label{sec:moments}

\begin{figure}[t]
\centering
\includegraphics[width=0.84\columnwidth]{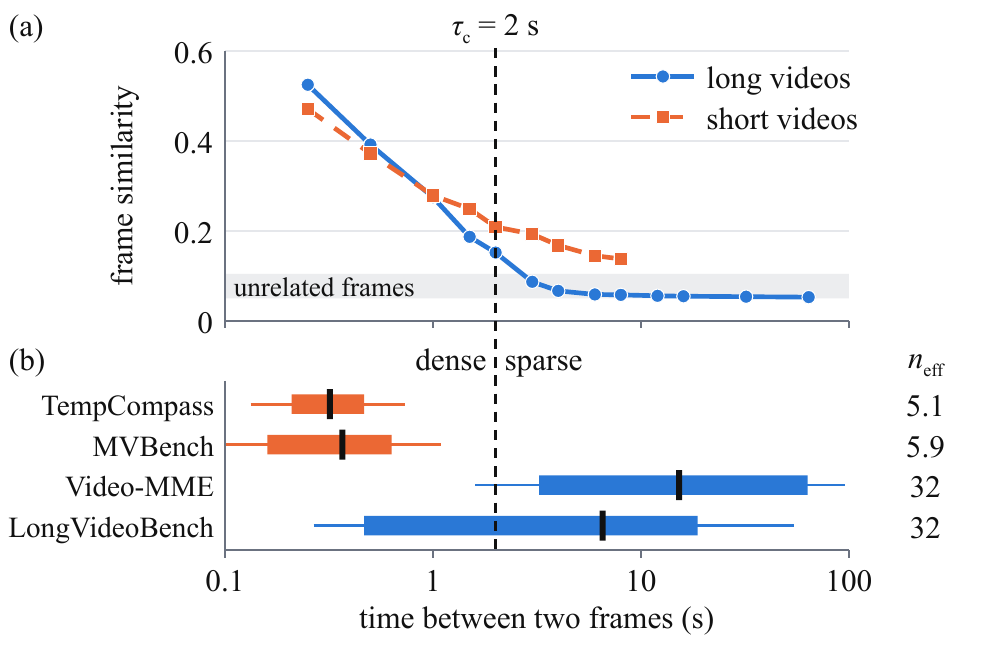}
\caption{Frames are not moments. (a) Phase-correlation peak between two frames of the same video against their temporal gap (median over videos). In long videos it reaches the range of unrelated frames (gray) between 2 and 3\,s. (b) Sampling interval of 32 uniform frames per benchmark (black: median; bar: quartiles; line: 5th--95th percentile) and the median number of moments ($\neff$, \cref{sec:count}).}
\label{fig:moments}
\end{figure}

We first measure how much new information a frame adds to its neighbors using the phase-correlation peak~\cite{kuglin1975phase}, which is robust to brightness shifts and global translation.
To separate temporal gap from video type, we vary the gap within each video from 0.25\,s to 64\,s (up to 8\,s on short clips), across 400 long videos (Video-MME, LongVideoBench) and 380 short clips (TempCompass, MVBench).
Pairs from different videos establish the unrelated baseline (median 0.050, 99th percentile 0.105).

In \cref{fig:moments}a, median similarity in long videos drops from 0.28 at 1\,s to 0.15 at 2\,s and 0.09 at 3\,s, and reaches the unrelated range between 2\,s and 3\,s.
We take this interval as the \emph{coherence time} of the video~\cite{goodman2015statistical}, the time over which a signal remains correlated with itself, and set $\tauc=2$\,s for all videos; frames less than $\tauc$ apart belong to one \emph{moment}.
Short clips stay correlated longer (0.14 at 8\,s) but are sampled far more densely than $\tauc$.
Any video longer than $T\tauc$ (about 1\,min at 32 frames, 8.5\,min at 256) is sampled more sparsely than $\tauc$; at 32 frames, the median video contains 5.1 and 5.9 moments on TempCompass and MVBench, but 32 on Video-MME and LongVideoBench (\cref{fig:moments}b).

\smallskip\noindent\textbf{Observation 1.} \emph{Frames decorrelate within about 2\,s. Under a token budget, a long video has as many moments as frames, and its frames are no longer temporally redundant.}

\subsection{Two Classes of Tokenizers}
\label{sec:rigid}

\begin{table}[t]
\centering
\scriptsize
\setlength{\tabcolsep}{2.6pt}
\caption{Tokenizer classes from model checkpoints. 3D-patch: two frames per patch ($u=2$), $\eta_t\ll1$, variable input size. Fixed-size: one frame per patch ($u=1$), one input resolution. Grid: learned position grid (--: none). $^*$Main; $^\circ$extension (\cref{sec:extension}).}
\label{tab:types}
\begin{tabular}{@{}lcc@{\hspace{7pt}}lcc@{}}
\toprule
\multicolumn{3}{@{}l}{3D-patch} & \multicolumn{3}{l@{}}{Fixed-size} \\
\cmidrule(r){1-3}\cmidrule(l){4-6}
Model & $\eta_t$ & Grid & Model & Grid & Input \\
\midrule
Qwen3-VL-8B$^*$ & 0.035 & $48^2$ & InternVL3.5-8B$^*$ & $32^2$ & 448 \\
Qwen3-Omni-30B & 0.055 & $48^2$ & InternVL3-8B & $32^2$ & 448 \\
Qwen2.5-VL-7B$^*$ & 0.042 & -- & InternVL2.5-8B & $32^2$ & 448 \\
Qwen2.5-Omni-7B & 0.005 & -- & LLaVA-OneVision-7B$^\circ$ & $27^2$ & 384 \\
Qwen2-VL-7B & 0.093 & -- & LLaVA-NeXT-Video-7B & $24^2$ & 336 \\
MiMo-VL-7B & 0.046 & -- & LLaVA-1.5-7B & $24^2$ & 336 \\
GLM-4.5V & 0.008 & $24^2$ & SmolVLM2-2.2B & $27^2$ & 384 \\
GLM-4.1V-9B$^\circ$ & 0.008 & $24^2$ & Granite-Vision-3.2-2B & $27^2$ & 384 \\
\bottomrule
\end{tabular}
\end{table}

Next, we inspect how tokenizers process these frames.
Across 16 open LLM checkpoints, we find two classes (\cref{tab:types}): \emph{3D-patch} tokenizers ($u=2$, variable input resolution) and \emph{fixed-size} tokenizers ($u=1$, single input resolution).
Aria fits neither class, and five models could not be classified (Appendix~B.2).

\begin{figure}[t]
\centering
\includegraphics[width=0.95\columnwidth]{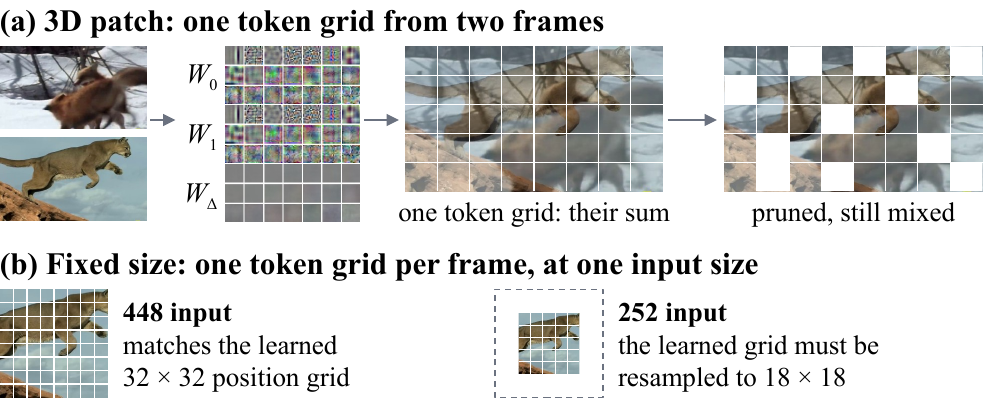}
\caption{Two classes of tokenizers. (a) The two temporal kernels of a 3D patch are nearly identical: 21 of the 1152 filters of Qwen3-VL-8B from the checkpoint, with $\Wdiff$ nearly flat on the same scale (over all filters, $\etat=0.035$). The token grid encodes the sum of the frames. (b) A fixed-size tokenizer encodes each frame independently, but with a single learned position grid.}
\label{fig:tokenizers}
\end{figure}

\noindent\textbf{3D-patch tokenizers.}
For $u=2$, the patch embedding applies temporal kernels $W_0$ and $W_1$.
Letting $\Wsum=W_0+W_1$ and $\Wdiff=W_0-W_1$, any frame pair decomposes as
\begin{equation}
  U(x_a,x_b)=\tfrac12\,\Wsum*(x_a+x_b)+\tfrac12\,\Wdiff*(x_a-x_b).
  \label{eq:decomp}
\end{equation}
The first term captures the frame sum and the second their difference; $\etat=\|\Wdiff\|/\|\Wsum\|$ quantifies sensitivity to the difference.
All eight 3D-patch models have $\etat\in[0.005,0.093]$, with Qwen3-VL's two kernels nearly identical (\cref{fig:tokenizers}a).
Since $W_0\approx W_1$, \cref{eq:decomp} simplifies to
\begin{equation}
  U(x_a,x_b)\approx\tfrac12\,\Wsum * (x_a+x_b),
  \quad U(x,x)=\Wsum * x ,
  \label{eq:fold}
\end{equation}
so each token grid encodes the sum of its two frames.

Summing nearby frames usually costs little, as they share similar content, but under sparse sampling each token superimposes two unrelated scenes, an effect we call \emph{double exposure} (\cref{fig:intro}a).
If the model used the difference between the two frames, swapping them or replacing both by their pixel mean would change its predictions; on videos sampled more than 2\,s apart, both keep accuracy within 0.6 points of the full model on Qwen3-VL and GLM-4.1V~\cite{glm2025v} (none significant).
In contrast, encoding every frame independently at full resolution, at twice the token count, improves accuracy by 2.4, 2.8 and 3.8 points on Qwen3-VL, Qwen2.5-VL and GLM-4.1V ($p<0.001$), and, at a fixed token count, separating pairs into stills still adds 1.15 points on Qwen3-VL ($p=0.011$, \cref{sec:length}).

\smallskip\noindent\textbf{Observation 2.} \emph{A 3D patch sums its two frames. Under sparse sampling, each token grid blends two moments, which limits the accuracy of the full model itself.}

\smallskip\noindent\textbf{Fixed-size tokenizers.}
Fixed-size tokenizers encode each frame independently, but at a single input resolution: that of InternVL matches its learned position grid ($32\times14=448$), so a smaller input requires resampling this grid (\cref{fig:tokenizers}b).
On densely sampled videos at the same budget, the two classes call for opposite reductions (\cref{tab:opposite}).
On InternVL, lowering the input resolution costs 2.4 points more than removing frames, and pooling the encoder output 1.6 points more; on Qwen-VL, lowering the resolution performs at least as well as removing frames, and removing frames brings no gain even at a higher resolution per grid.

\smallskip\noindent\textbf{Observation 3.} \emph{The least costly reduction depends on the tokenizer. On densely sampled videos, a 3D-patch tokenizer should lower its spatial resolution, whereas a fixed-size tokenizer should retain its input resolution and remove frames.}

\subsection{Implications}
\label{sec:failures}

These findings point to a limit on long videos that token selection does not address.
Token-space methods keep, discard or merge the tokens of $Z$.
On a 3D-patch tokenizer, however, the blend arises in the patch embedding: two pairs with the same sum receive nearly the same embedding, since the kernel norm of $\Wdiff$ is only a fraction $\etat$ of that of $\Wsum$, and the model gains little from this term on sparse videos (\cref{sec:rigid}).
Pruning or merging can therefore only retain, discard or average blended tokens, and uniform downscaling reduces $b$ without changing which frames share a grid.
The critical decisions precede tokenization: the number of grids, the dimension to reduce, and the stage at which to reduce it.

\section{ESSENCE}
\label{sec:method}

\begin{figure*}[t]
\centering
\includegraphics[width=0.98\textwidth]{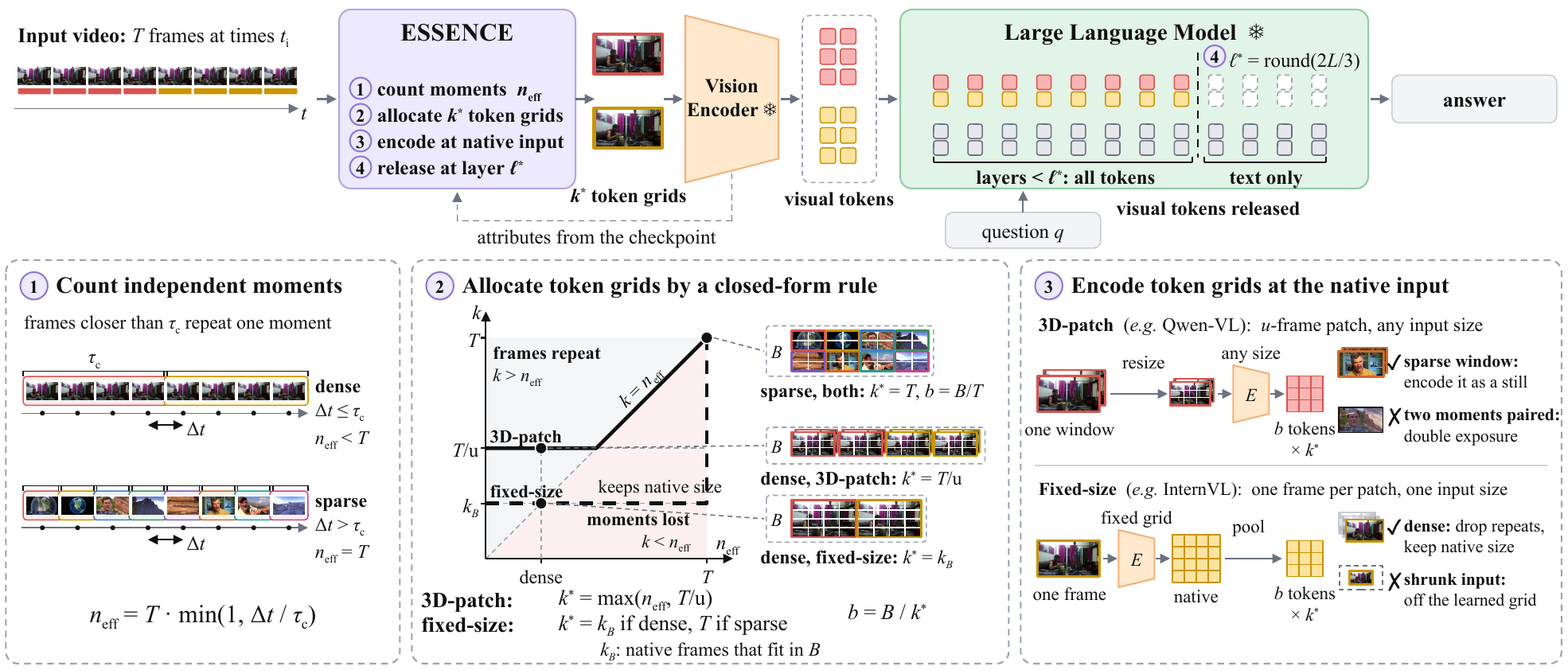}
\caption{Overview of ESSENCE. Each color denotes one moment. (1) Frames separated by less than $\tauc$ belong to one moment. (2) A closed-form rule sets the number of token grids, each with at most $B/\kstar$ tokens (dashed boxes: the same budget $B$). (3) Each grid is encoded without modifying the model: Qwen-VL resizes its input, and InternVL pools its encoder output on sparse videos. (4) Visual tokens are released from the LLM at layer $\ell^*$.}
\label{fig:method}
\end{figure*}

Every visual token incurs computation from the vision encoder through the final LLM layer.
Using only frame timestamps and the model checkpoint, ESSENCE decides before encoding which tokens to form and where to reduce them, guided by two principles.
First, \emph{retain a visual token only while it can affect the prediction}: tokens from the same moment are often redundant, and visual tokens in late LLM layers rarely alter the output.
Second, \emph{apply each reduction at the earliest stage that preserves accuracy}, which then saves computation in all later stages.
ESSENCE needs no training and no labels at inference, and shares its two constants, $\tauc$ and the release ratio, across all videos and models.

\subsection{Count Independent Moments}
\label{sec:count}

Because a token helps only when it adds non-redundant information, step one estimates the number of moments in a video.
With frames closer than $\tauc$ grouped into one moment, a video of duration $D=T\Delta t$ spans about $D/\tauc$ moments, capped by the frame count:
\begin{equation}
  \neff = T\cdot\min(1,\ \Delta t/\tauc) = \min(T,\ D/\tauc).
  \label{eq:neff}
\end{equation}
Like the effective sample size of a correlated series~\cite{geyer1992practical}, $\neff$ estimates the number of non-redundant frames; it is a proxy for coverage, not a count of semantic events.
We call a video \emph{dense} if $\Delta t\le\tauc$ and \emph{sparse} otherwise; in a sparse video, every frame is a separate moment.
Unlike token-similarity metrics, $\neff$ uses only timestamps and is available before tokenization.
Content-based estimates add at most 0.07 points, and halving or doubling $\tauc$ shifts accuracy by at most 0.6 points (Appendix~C.2--C.3).

\subsection{Allocate Token Grids by a Closed-Form Rule}
\label{sec:allocate}

Step two splits the budget between the grid count $k$ and tokens per grid $b$.
Without labels to optimize this split per video, we derive it from the reduction costs in Observation~3.
A smaller grid only reduces spatial detail within a moment, whereas omitting a grid loses that moment entirely; thus every moment receives at least one grid.
The least costly reduction of the tokenizer leaves $k_0$ grids, and no video receives fewer.
Combining both conditions gives the closed-form rule (\cref{fig:method}, step 2):
\begin{equation}
  \kstar=\max(\neff,\ k_0),
  \label{eq:rule}
\end{equation}
with $\kstar$ rounded to an integer and $b=\min(b_{\text{native}},B/\kstar)$ rounded down to a valid grid size.

For 3D-patch tokenizers, the encoder accepts lower resolutions, so $k_0=T/u$: the model starts with its native grid count, adds grids once moments outnumber them (up to one per frame), and splits the budget equally across grids.
At $T=32$ and $u=2$, a 16\,s clip uses 16 native grids, a 48\,s video uses 24, and a video of 64\,s or longer uses 32 (one per frame); in every case, paired frames are at most $\tauc$ apart.

For fixed-size tokenizers, lowering input resolution on dense videos hurts accuracy more than removing redundant frames (\cref{tab:opposite}), so $k_0=k_B=\lfloor B/b_{\text{native}}\rfloor$, the number of native-resolution frames fitting within $B$.
For these tokenizers, we apply \cref{eq:rule} as a step: $\kstar=T$ on sparse videos and $\kstar=k_0$ on dense ones. Following \cref{eq:rule} between the two, i.e., pooling $\mathrm{round}(\neff)$ frames when $\neff>k_B$, gives nearly the same predictions at a higher cost (Appendix~C.4).

\subsection{Encode Token Grids at the Native Input}
\label{sec:realize}

Step three realizes this allocation without changing model weights.
A 3D-patch tokenizer expects frame pairs, whereas producing more than $T/u$ grids requires encoding individual frames.
Native processors already support this: Qwen-VL encodes a still image by duplicating it, and by \cref{eq:fold} the resulting grid represents that frame alone.
We partition the $T$ frames into $\kstar$ consecutive windows of nearly equal duration, encoding two-frame windows as native pairs and one-frame windows as duplicated stills while preserving original timestamps (Appendix~A.1).
Sparse videos ($\kstar=T$) are encoded entirely as stills, one per moment, without double exposure.

The second principle determines where each grid is reduced to $b$ tokens.
Because 3D-patch encoders support variable resolutions, we resize each input before patch embedding, which also reduces the encoding latency.
Fixed-size encoders operate at the resolution of their learned position grid, so we encode the retained frames at native resolution and average-pool their encoder outputs to $b$ tokens (on dense videos, $b=b_{\text{native}}$ requires no pooling).

\subsection{Release Visual Tokens at Layer \texorpdfstring{$\ell^*$}{l*}}
\label{sec:release}

\begin{figure}[t]
\centering
\includegraphics[width=0.58\columnwidth]{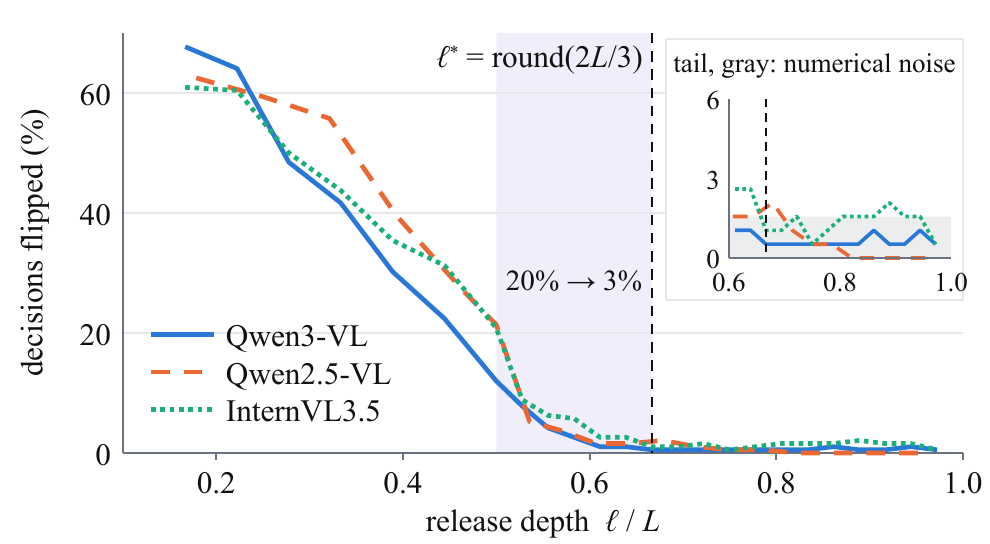}
\caption{Late layers rarely change the prediction. Fraction of first-token decisions that change when visual tokens are released from layer $\ell$, on 192 unlabeled calibration questions; after $2L/3$ it stays below 3\%. Inset: layers past $0.6L$; gray: flips from numerical noise alone (an equivalent attention kernel).}
\label{fig:horizon}
\end{figure}

Within the LLM, prefill processes all visual tokens at every layer, yet by our first principle they are needed only while they influence the output.
Let $f(\ell)$ be the fraction of first-token predictions that flip when visual tokens are removed from layer $\ell$ onward (zero-based); it requires no labels, and we measure it on 192 calibration questions from 64 videos outside the benchmarks.
On all three models, $f$ drops sharply past $0.5L$ and stays below 3\% after $0.67L$ (\cref{fig:horizon}).
ESSENCE therefore releases visual tokens at $\ell^*=\mathrm{round}(2L/3)$, one relative depth across models rather than the per-model layer of VTW~\cite{lin2025boosting}.
From this layer onward, visual tokens are removed from the hidden states and neither updated nor cached; earlier layers retain their visual keys and values, which decoding still uses.
For multiple-choice questions the first token is the answer, so on any set of questions the fraction of flipped predictions bounds the accuracy change.
At $\ell^*$, $f$ is 0.5\%, 2.1\% and 1.0\% on Qwen3-VL, Qwen2.5-VL and InternVL3.5 (Appendix~D.1).

\setcounter{dbltopnumber}{3}
\setcounter{topnumber}{3}
\setcounter{totalnumber}{6}
\renewcommand{\dbltopfraction}{0.95}
\renewcommand{\topfraction}{0.95}
\renewcommand{\textfraction}{0.02}

\section{Experiments}
\label{sec:experiments}

\begin{table*}[t]
\centering
\scriptsize
\setlength{\tabcolsep}{8pt}
\renewcommand{\arraystretch}{.82}
\caption{Accuracy, TTFT and E2E latency under one timing protocol, from decoded frames to the first answer token and to 32 generated tokens, with speedups over the full model. ESSENCE-$x$: $x$\% of the full model's tokens. Best in bold, second underlined, excluding the full model. $^\dagger$Authors' modified Transformers, 1.44$\times$ slower than the native stack (Appendix~A.4). $^\ddagger$Qwen-VL only (Appendix~E.3).}
\label{tab:main}
\begin{tabular}{@{}lccccrrrrr@{}}
\toprule
 & \multicolumn{5}{c}{Accuracy (\%)} & \multicolumn{2}{c}{TTFT} & \multicolumn{2}{c}{E2E} \\
\cmidrule(lr){2-6}\cmidrule(lr){7-8}\cmidrule(l){9-10}
Method & TempCompass & Video-MME & MVBench & LongVideoBench & Avg & ms & Speedup & ms & Speedup \\
\midrule
\multicolumn{10}{l}{\textit{Qwen3-VL-8B}} \\
Full model & 74.49 & 64.04 & 68.76 & 59.91 & 66.80 & 401 & 1.00$\times$ & 1556 & 1.00$\times$ \\
FastVID\raisebox{-.4ex}[0pt][0pt]{\tiny\,NeurIPS'25} & 69.94 & 58.93 & 65.97 & 57.52 & 63.09 & 237 & 1.69$\times$ & 1392 & 1.12$\times$ \\
FlashVID\raisebox{-.4ex}[0pt][0pt]{\tiny\,ICLR'26} & 71.90 & 60.74 & 66.74 & 58.86 & 64.56 & 255 & 1.57$\times$ & 1413 & 1.10$\times$ \\
iLLaVA\raisebox{-.4ex}[0pt][0pt]{\tiny\,ICLR'26} & 72.72 & 61.26 & 67.05 & 59.84 & 65.22 & 619 & 0.65$\times$$^\dagger$ & 1757 & 0.89$\times$ \\
EarlyTom\raisebox{-.4ex}[0pt][0pt]{\tiny\,CVPR'26} & 71.77 & 60.85 & 66.24 & 55.87 & 63.68 & 226 & 1.77$\times$ & 1387 & 1.12$\times$ \\
ShortV\raisebox{-.4ex}[0pt][0pt]{\tiny\,ICCV'25} & 72.59 & 60.67 & 67.55 & \underline{61.26} & 65.52 & 265 & 1.51$\times$ & 1418 & 1.10$\times$ \\
Q-Frame$^\ddagger$\raisebox{-.4ex}[0pt][0pt]{\tiny\,ICCV'25} & 65.38 & 62.11 & 62.13 & 59.54 & 62.29 & 732 & 0.55$\times$ & 1879 & 0.83$\times$ \\
LDDR$^\ddagger$\raisebox{-.4ex}[0pt][0pt]{\tiny\,arXiv'26} & 69.05 & 64.22 & 63.89 & 60.73 & 64.48 & 774 & 0.52$\times$ & 1925 & 0.82$\times$ \\
\cmidrule(lr){1-10}
ESSENCE-80 & \textbf{74.18} & \textbf{64.74} & \textbf{68.76} & \textbf{61.56} & \textbf{67.31} & 248 & 1.62$\times$ & 1398 & 1.11$\times$ \\
ESSENCE-66 & \textbf{74.18} & \underline{64.48} & \underline{68.00} & 60.73 & \underline{66.85} & \underline{225} & \underline{1.78$\times$} & \underline{1375} & \underline{1.13$\times$} \\
ESSENCE-42 & \underline{73.80} & 62.26 & 66.89 & 58.71 & 65.42 & \textbf{172} & \textbf{2.32$\times$} & \textbf{1320} & \textbf{1.18$\times$} \\
\midrule
\multicolumn{10}{l}{\textit{Qwen2.5-VL-7B}} \\
Full model & 71.84 & 59.67 & 64.26 & 55.65 & 62.85 & 503 & 1.00$\times$ & 1343 & 1.00$\times$ \\
FastVID\raisebox{-.4ex}[0pt][0pt]{\tiny\,NeurIPS'25} & 68.80 & 55.85 & 61.05 & 54.08 & 59.94 & 327 & 1.54$\times$ & 1159 & 1.16$\times$ \\
FlashVID\raisebox{-.4ex}[0pt][0pt]{\tiny\,ICLR'26} & 69.11 & 57.56 & 62.53 & 54.60 & 60.95 & 352 & 1.43$\times$ & 1183 & 1.14$\times$ \\
iLLaVA\raisebox{-.4ex}[0pt][0pt]{\tiny\,ICLR'26} & \underline{70.82} & 57.89 & 62.18 & 53.85 & 61.19 & 322 & 1.56$\times$ & 1163 & 1.15$\times$ \\
EarlyTom\raisebox{-.4ex}[0pt][0pt]{\tiny\,CVPR'26} & 68.92 & 56.52 & 61.16 & 53.78 & 60.09 & 272 & 1.85$\times$ & 1106 & 1.22$\times$ \\
ShortV\raisebox{-.4ex}[0pt][0pt]{\tiny\,ICCV'25} & 67.59 & 55.78 & 60.00 & 53.78 & 59.29 & 365 & 1.38$\times$ & 1191 & 1.13$\times$ \\
Q-Frame$^\ddagger$\raisebox{-.4ex}[0pt][0pt]{\tiny\,ICCV'25} & 68.04 & 57.74 & 59.11 & \textbf{57.37} & 60.56 & 798 & 0.63$\times$ & 1627 & 0.82$\times$ \\
LDDR$^\ddagger$\raisebox{-.4ex}[0pt][0pt]{\tiny\,arXiv'26} & 69.87 & 59.04 & 60.87 & 55.87 & 61.41 & 834 & 0.61$\times$ & 1659 & 0.81$\times$ \\
\cmidrule(lr){1-10}
ESSENCE-80 & \textbf{71.52} & \textbf{60.78} & \underline{62.66} & \underline{56.77} & \textbf{62.93} & 317 & 1.59$\times$ & 1147 & 1.17$\times$ \\
ESSENCE-66 & 70.38 & \underline{60.00} & \textbf{63.11} & 55.42 & \underline{62.23} & \underline{265} & \underline{1.90$\times$} & \underline{1095} & \underline{1.23$\times$} \\
ESSENCE-42 & 69.49 & 59.30 & 62.03 & 54.82 & 61.41 & \textbf{177} & \textbf{2.84$\times$} & \textbf{1009} & \textbf{1.33$\times$} \\
\midrule
\multicolumn{10}{l}{\textit{InternVL3.5-8B}} \\
Full model & 71.14 & 64.74 & 73.39 & 62.53 & 67.95 & 1323 & 1.00$\times$ & 2452 & 1.00$\times$ \\
FastVID\raisebox{-.4ex}[0pt][0pt]{\tiny\,NeurIPS'25} & 68.54 & 61.63 & 67.82 & 60.36 & 64.59 & 732 & 1.81$\times$ & 1850 & 1.33$\times$ \\
FlashVID\raisebox{-.4ex}[0pt][0pt]{\tiny\,ICLR'26} & 68.61 & 62.41 & 66.76 & \underline{60.73} & 64.63 & 846 & 1.57$\times$ & 1969 & 1.25$\times$ \\
iLLaVA\raisebox{-.4ex}[0pt][0pt]{\tiny\,ICLR'26} & \underline{70.44} & \underline{63.11} & 71.16 & \textbf{61.11} & \underline{66.45} & 1267 & 1.04$\times$ & 2389 & 1.03$\times$ \\
EarlyTom\raisebox{-.4ex}[0pt][0pt]{\tiny\,CVPR'26} & 69.30 & 60.81 & 68.92 & 58.12 & 64.29 & 751 & 1.76$\times$ & 1903 & 1.29$\times$ \\
ShortV\raisebox{-.4ex}[0pt][0pt]{\tiny\,ICCV'25} & 65.89 & 57.19 & 65.42 & 58.86 & 61.84 & 907 & 1.46$\times$ & 2040 & 1.20$\times$ \\
\cmidrule(lr){1-10}
ESSENCE-56 & \textbf{70.63} & \textbf{64.33} & \textbf{72.58} & \underline{60.73} & \textbf{67.07} & \underline{678} & \underline{1.95$\times$} & \underline{1815} & \underline{1.35$\times$} \\
ESSENCE-32 & \underline{70.44} & 62.48 & \underline{72.55} & 60.13 & 66.40 & \textbf{485} & \textbf{2.72$\times$} & \textbf{1610} & \textbf{1.52$\times$} \\
\bottomrule
\end{tabular}
\end{table*}

\begin{table}[t]
\centering
\scriptsize
\setlength{\tabcolsep}{1.5pt}
\caption{Accuracy gain of ESSENCE over each baseline at matched mean TTFT, in points. $^*$: $p<0.05$. $^\dagger$: ESSENCE is slower on this benchmark alone, by at most 10.6\%.}
\label{tab:matched}
\begin{tabular}{@{}l*{4}{c}@{\hspace{4pt}}*{4}{c}@{\hspace{4pt}}*{4}{c}@{}}
\toprule
 & \multicolumn{4}{c}{Qwen3-VL} & \multicolumn{4}{c}{Qwen2.5-VL} & \multicolumn{4}{c}{InternVL} \\
\cmidrule(lr){2-5}\cmidrule(lr){6-9}\cmidrule(l){10-13}
 & TC & VM & MV & LVB & TC & VM & MV & LVB & TC & VM & MV & LVB \\
\midrule
FastVID & \cellcolor{blue!22}4.2$^*$ & \cellcolor{blue!29}5.6$^*$ & \cellcolor{blue!11}2.0$^*$ & \cellcolor{blue!17}3.2$^*$ & \cellcolor{blue!14}2.7$^{*\dagger}$ & \cellcolor{blue!26}4.9$^*$ & \cellcolor{blue!8}1.6$^{*\dagger}$ & \cellcolor{blue!14}2.7$^*$ & \cellcolor{blue!11}2.1$^*$ & \cellcolor{blue!14}2.7$^{*\dagger}$ & \cellcolor{blue!25}4.8$^*$ & \cellcolor{blue!2}0.4 \\
FlashVID & \cellcolor{blue!12}2.3$^*$ & \cellcolor{blue!21}4.0$^*$ & \cellcolor{blue!11}2.0$^{*\dagger}$ & \cellcolor{blue!14}2.7$^*$ & \cellcolor{blue!13}2.4$^*$ & \cellcolor{blue!17}3.2$^*$ & 0.1$^\dagger$ & \cellcolor{blue!11}2.2$^*$ & \cellcolor{blue!11}2.0$^*$ & \cellcolor{blue!10}1.9$^*$ & \cellcolor{blue!31}5.8$^*$ & 0.0 \\
iLLaVA & \cellcolor{blue!8}1.5 & \cellcolor{blue!18}3.5$^*$ & \cellcolor{blue!9}1.7$^*$ & \cellcolor{blue!9}1.7 & \cellcolor{blue!4}0.7 & \cellcolor{blue!15}2.9$^*$ & \cellcolor{blue!2}0.5$^\dagger$ & \cellcolor{blue!15}2.9$^*$ & 0.2 & \cellcolor{blue!6}1.2$^*$ & \cellcolor{blue!7}1.4$^*$ & $-$0.4 \\
EarlyTom & \cellcolor{blue!13}2.4$^*$ & \cellcolor{blue!19}3.6$^*$ & \cellcolor{blue!9}1.8$^{*\dagger}$ & \cellcolor{blue!26}4.9$^*$ & \cellcolor{blue!8}1.5 & \cellcolor{blue!18}3.5$^*$ & \cellcolor{blue!10}1.9$^{*\dagger}$ & \cellcolor{blue!9}1.6 & \cellcolor{blue!7}1.3$^*$ & \cellcolor{blue!19}3.5$^{*\dagger}$ & \cellcolor{blue!19}3.7$^*$ & \cellcolor{blue!14}2.6$^*$ \\
ShortV & \cellcolor{blue!8}1.6 & \cellcolor{blue!21}4.1$^*$ & \cellcolor{blue!6}1.2$^{*\dagger}$ & \cellcolor{blue!2}0.3 & \cellcolor{blue!21}3.9$^*$ & \cellcolor{blue!26}5.0$^*$ & \cellcolor{blue!14}2.7$^*$ & \cellcolor{blue!16}3.0$^*$ & \cellcolor{blue!25}4.7$^*$ & \cellcolor{blue!38}7.1$^*$ & \cellcolor{blue!38}7.2$^*$ & \cellcolor{blue!10}1.9 \\
\bottomrule
\end{tabular}
\end{table}

\begin{figure*}[t]
\centering
\includegraphics[width=0.79\textwidth]{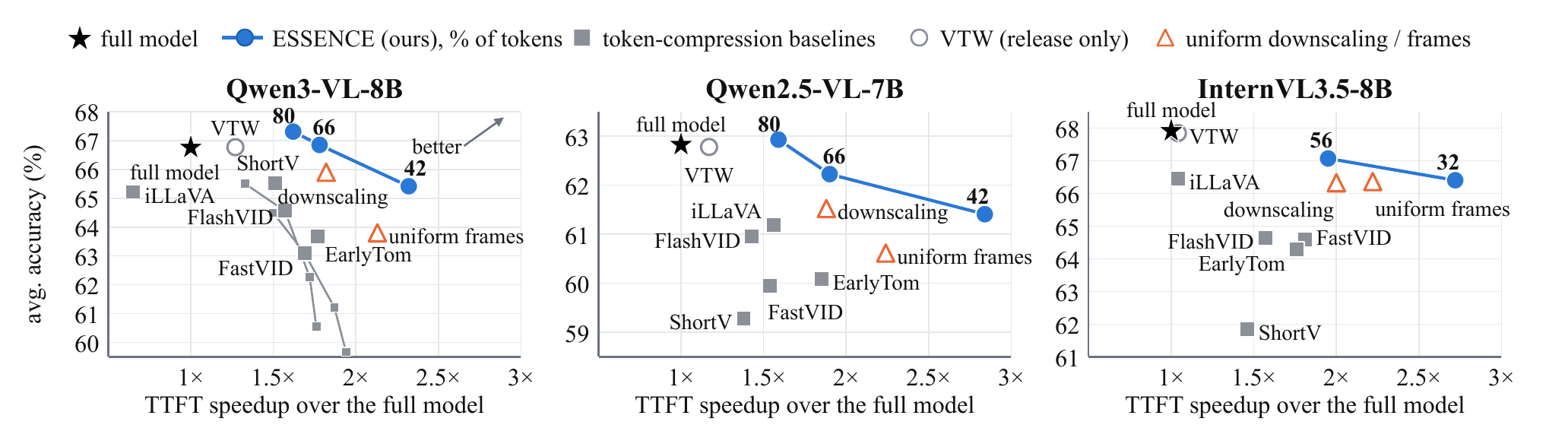}
\caption{Average accuracy against TTFT speedup; numbers: ESSENCE budget (\%). Downscaling: budget of ESSENCE-66 (56 on InternVL). Uniform frames: 16 (Qwen-VL) or 18 (InternVL) frames at native resolution. VTW: all tokens, released at the layer chosen by VTW. Thin lines: FastVID and FlashVID at retention .10, .15, default and .40.}
\label{fig:frontier}
\end{figure*}

\begin{figure*}[t]
\centering
\includegraphics[width=0.80\textwidth]{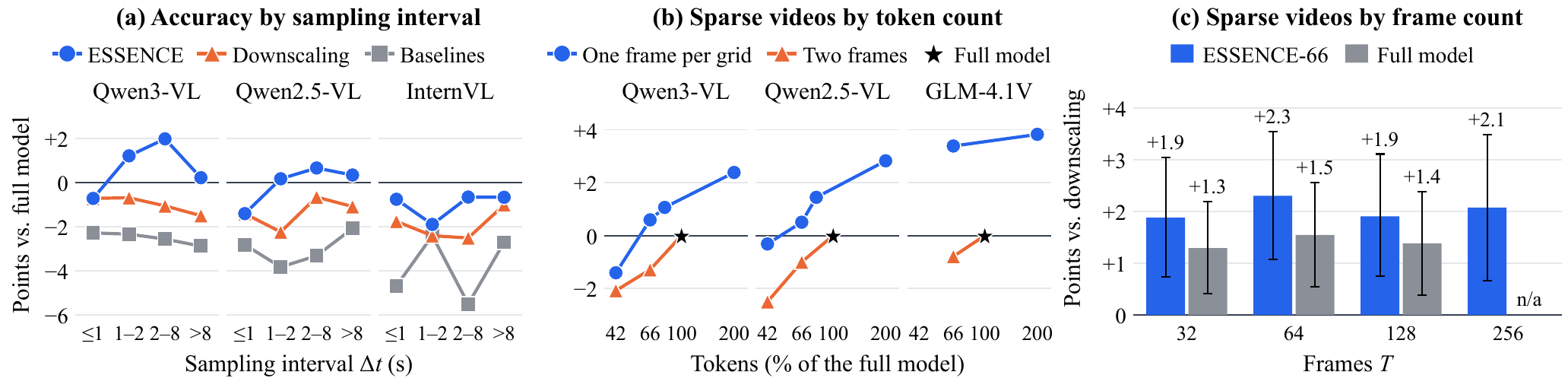}
\caption{Source of the improvement at the 66\% budget (56\% on InternVL). (a) Accuracy by sampling interval; baselines: mean of five. (b) Sparse videos of VM and LVB: one- vs.\ two-frame grids by token count (100\%: full model; 200\%: every frame at full size). (c) Lead over downscaling on these videos on Qwen3-VL (95\% CIs); the budget scales with $T$; the full model is omitted at 256 frames (about 32k tokens).}
\label{fig:analysis}
\end{figure*}

\begin{figure}[t]
\centering
\includegraphics[width=0.84\columnwidth]{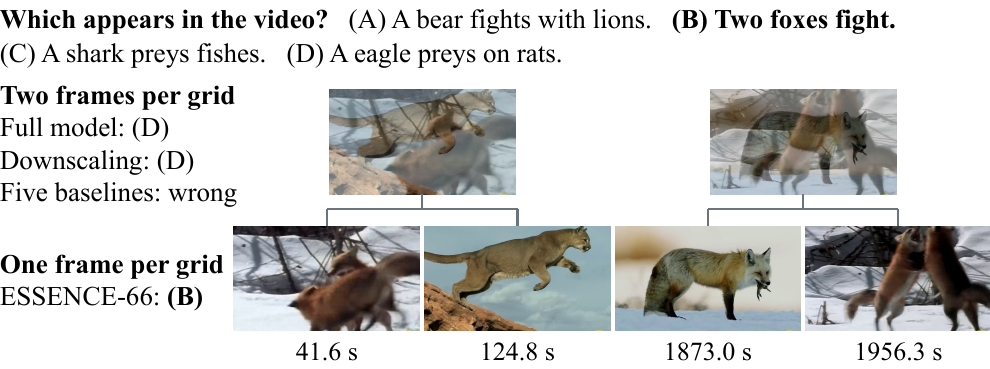}
\caption{A selected example from Video-MME on Qwen3-VL, with frames 83\,s apart. Pairing blends each fox scene with its neighboring frame, and the full model, downscaling and all five baselines answer incorrectly; ESSENCE-66 encodes one frame per grid and answers correctly.}
\label{fig:case}
\end{figure}

\begin{table}[t]
\centering
\scriptsize
\setlength{\tabcolsep}{3.2pt}
\caption{Gain of the rule over two alternatives at a similar token count and the smallest budget (42\%, InternVL 32\%). Pair/still: two/one frames per grid; sizes in pixels; 8 pairs and 10 frames drop frames, others downscale or pool. $^\dagger$: 17--32 pairs/stills at $1<\Delta t\le2$\,s. $^\circ$: TC, MV only; $^\diamond$: VM, LVB only. $^*$: $p<0.05$.}
\label{tab:opposite}
\begin{tabular}{@{}llrrr@{}}
\toprule
 & & \multicolumn{2}{c}{3D-patch} & Fixed-size \\
\cmidrule(lr){3-4}\cmidrule(l){5-5}
Rule & Against & Qwen3 & Qwen2.5 & InternVL \\
\midrule
\multicolumn{5}{@{}l}{\textit{Dense videos}} \\
16 pairs of $292^2$$^\dagger$ & 8 pairs of $413^2$ & $+0.72$ & $+0.23$ & \\
 & 32 stills of $206^2$$^\circ$ & $+1.88^*$ & $+1.42^*$ & \\
10 frames of 448 & 32 frames, pooled$^\circ$ & & & $+1.57^*$ \\
 & 32 frames of 252 & & & $+2.39^*$ \\
\midrule
\multicolumn{5}{@{}l}{\textit{Sparse videos}} \\
32 stills of $206^2$ & 16 pairs of $292^2$ & $+0.70$ & $+2.17^{*\diamond}$ & \\
 & 8 pairs of $413^2$ & $+2.92^*$ & $+2.98^*$ & \\
32 frames, pooled & 10 frames of 448$^\diamond$ & & & $+2.41^*$ \\
 & 32 frames of 252 & & & $+1.23^*$ \\
\bottomrule
\end{tabular}
\end{table}

\begin{table}[t]
\centering
\scriptsize
\setlength{\tabcolsep}{1.2pt}
\caption{Ablation at 66\% of the tokens (56\% on InternVL). Accuracy over all 9417 questions and the 3425 sparse ones (per question, unlike Avg in \cref{tab:main}). W/o allocation: uniform downscaling to the same budget. $^*$: below ESSENCE, $p<0.05$.}
\label{tab:ablation}
\begin{tabular}{@{}l*{9}{r}@{}}
\toprule
 & \multicolumn{3}{c}{Qwen3-VL} & \multicolumn{3}{c}{Qwen2.5-VL} & \multicolumn{3}{c}{InternVL} \\
\cmidrule(lr){2-4}\cmidrule(lr){5-7}\cmidrule(l){8-10}
Variant & All & Sparse & TTFT & All & Sparse & TTFT & All & Sparse & TTFT \\
\midrule
Full model & 67.11 & 60.15 & 1.00$\times$ & 62.99 & 56.06 & 1.00$\times$ & 68.99 & 61.58 & 1.00$\times$ \\
w/o release & 66.85 & 60.55 & 1.44$\times$ & 62.43 & 56.35 & 1.59$\times$ & 68.29 & 60.70 & 1.64$\times$ \\
w/o allocation & 66.15$^*$ & 58.74$^*$ & 1.82$\times$ & 61.69$^*$ & 55.07$^*$ & 1.88$\times$ & 67.36$^*$ & 60.23 & 2.00$\times$ \\
\textbf{ESSENCE} & \textbf{67.00} & \textbf{60.76} & \textbf{1.78}$\times$ & \textbf{62.34} & \textbf{56.47} & \textbf{1.90}$\times$ & \textbf{68.21} & \textbf{60.91} & \textbf{1.95}$\times$ \\
\bottomrule
\end{tabular}
\end{table}

\begin{figure}[t]
\centering
\includegraphics[width=\columnwidth]{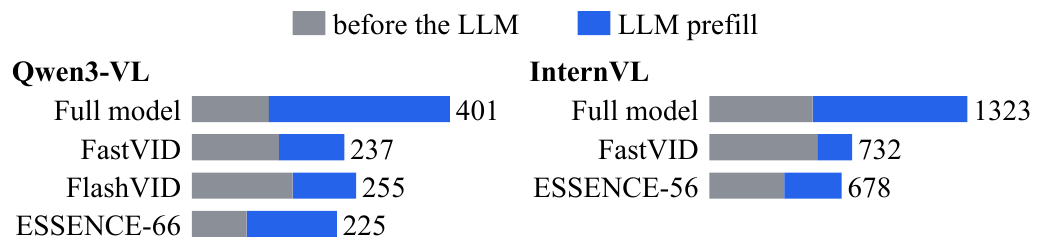}
\caption{TTFT by stage relative to each full model (totals in ms).}
\label{fig:stage}
\end{figure}

\subsection{Experimental Settings}
\label{sec:setup}

\noindent\textbf{Models and benchmarks.}
We evaluate two 3D-patch models, Qwen3-VL-8B~\cite{bai2025qwen3} and Qwen2.5-VL-7B~\cite{bai2025qwen25}, and one fixed-size model, InternVL3.5-8B~\cite{zhu2025internvl3}.
Four multiple-choice benchmarks span the sampling intervals in \cref{fig:moments}b: TempCompass (TC, 1580 questions)~\cite{liu2024tempcompass} and MVBench (MV, 3800)~\cite{li2024mvbench} consist mostly of short clips, whereas Video-MME (VM, 2700)~\cite{fu2024videomme} and LongVideoBench (LVB, 1337)~\cite{wu2024longvideobench} consist mostly of long videos.
Every method receives the same 32 uniformly sampled frames.

\noindent\textbf{Baselines.}
We compare against five training-free accelerators: post-encoder FastVID~\cite{shao2025fastvid} and FlashVID~\cite{cao2026flashvid}, in-encoder iLLaVA~\cite{hu2026illava} and EarlyTom~\cite{earlytom2026}, and in-LLM ShortV~\cite{yuan2025shortv}.
We also compare with uniform downscaling plus release, the low-resolution stream of concurrent LoHi~\cite{dong2026lohi} (code unreleased; Appendix~A.3).
On Qwen-VL, we also evaluate pre-encoder question-guided frame selectors Q-Frame~\cite{zhang2025qframe} and LDDR~\cite{lddr2026} (selection included in timing; Appendix~E.3).
For 12 of the 15 baseline-backbone pairs, we use our own ports (Appendix~A.3).

\noindent\textbf{Budgets and metrics.}
The token budget is the sole user parameter.
Accuracy differences are reported in percentage points (points) and tested via two-sided video-level permutation tests~\cite{ernst2004permutation}; we call a difference significant if $p<0.05$.
Latency is timed on one RTX 5090 over 192 stratified requests, starting from decoded frames in host memory to the first answer token (TTFT) and to 32 generated tokens (E2E).

\subsection{Comparison with the State of the Art}
\label{sec:sota}

\Cref{tab:main} shows that on each backbone, ESSENCE achieves the highest average accuracy, the fastest TTFT and the lowest E2E latency, each at one of its budgets.
On Qwen-VL, ESSENCE-80 retains 80\% of the tokens and matches the average accuracy of the full models ($+0.5$ and $+0.1$ points) at $1.6\times$ faster TTFT.
For a comparison at equal cost, we pair each baseline with the largest ESSENCE budget whose mean TTFT over the four benchmarks is at least as fast (\cref{tab:matched}).
\textbf{ESSENCE achieves higher accuracy in 58 of these 60 comparisons, 46 of them significantly, and is significantly worse in none.}
The 46 include all 15 comparisons on VM, and 38 of them lie in cells where ESSENCE is not slower even on that benchmark alone (no $^\dagger$); 27 of the 60 comparisons survive Bonferroni correction (Appendix~E.2).
On the speed-accuracy trade-off (\cref{fig:frontier}), each token-compression baseline is outperformed on both axes by at least one ESSENCE budget, and on Qwen3-VL ESSENCE Pareto-dominates FastVID and FlashVID at all four retention ratios.

\subsection{Effect of the Sampling Interval}
\label{sec:length}

\noindent\textbf{Accuracy by interval.}
Grouping the questions by $\Delta t$ (\cref{fig:analysis}a) shows where allocation takes effect.
Downscaling is itself a strong reference: it leads the five baselines in 51 of 60 comparisons (Appendix~E.2), and on Qwen-VL it receives the same input as ESSENCE when $\Delta t\le1$\,s.
Once $\Delta t$ exceeds 1\,s, ESSENCE diverges from downscaling and reaches the accuracy of the full model, confirming that allocation matters once frames decorrelate.
On sparse videos, all five baselines and downscaling fall significantly below the full model, whereas ESSENCE-66 and ESSENCE-56 do not, with 1.8--2.0$\times$ faster TTFT.
On hour-scale subsets (15--60\,min), ESSENCE stays above the full model on Qwen-VL but falls 3.0 points below it on LVB with InternVL, as does downscaling (3.2 points); on a fixed-size tokenizer, any reduction can only remove spatial detail (Appendix~E.4).

\noindent\textbf{Double exposure.}
To isolate double exposure, we compare one-frame and two-frame grids across token counts on sparse videos (\cref{fig:analysis}b).
One-frame grids achieve higher accuracy than two-frame grids at every measured token count on all three 3D-patch models, and reach the accuracy of the full model at 66\% of its tokens (\cref{fig:case} shows an example).
In a $2\times2$ design at a fixed token count on Qwen3-VL, 32 rather than 16 frames add 1.76 points and stills rather than pairs add 1.15 points, both significantly, so both the number of moments and double exposure limit accuracy (Appendix~B.3).

\noindent\textbf{Frame count.}
More frames do not remove sparsity: at 256 frames, 49\% of VM and LVB questions remain sparse.
As \cref{fig:analysis}c shows, ESSENCE-66 keeps a significant lead of about 2 points over downscaling up to 256 frames, and stays 0.8 and 0.5 points above the full model at 64 and 128 frames, not significantly, with 1.8--1.9$\times$ lower time per question.

\subsection{Effect of the Tokenizer}
\label{sec:opposite}

\Cref{tab:opposite} tests whether the least costly reduction reverses with the tokenizer class, as Observation~3 predicts.
The rule outperforms both alternatives in all 12 comparisons, 9 of them significantly.
As predicted, retaining all 32 frames is better on sparse videos for both classes, but on dense videos it is worse on InternVL only (interaction $p<0.01$).

\subsection{Ablation Study}
\label{sec:ablation}

\Cref{tab:ablation} removes allocation and release in turn.
Allocation drives the accuracy gain: without it, accuracy drops significantly on all three models at about the same speed, by 2.0 and 1.4 points on the sparse videos of Qwen3-VL and Qwen2.5-VL.
Release drives the remaining speedup: without it, the TTFT speedup falls from 1.78--1.95$\times$ to 1.44--1.64$\times$.
Release changes only 1.0\%, 3.7\% and 1.7\% of the predictions on the three models and no overall accuracy significantly (Appendix~D.2); for multi-turn dialogue, it can be disabled at the cost of this extra speedup (Appendix~D.4).

Stage-wise timing (\cref{fig:stage}) shows that ESSENCE reduces the latency before the LLM by 28--29\% and the prefill by 50--63\%, whereas FastVID and FlashVID run the full encoder.
With our release added, FlashVID and FastVID change by at most 0.1 points overall and 0.3 points on any benchmark, and ESSENCE-66 still leads them by 2.1 and 3.6 points on Qwen3-VL, both significantly, so the lead comes from allocation.
Still-frame encoding alone improves the VM accuracy of FastVID, FlashVID and ShortV (Appendix~E.7).

\subsection{Extension to Other Models}
\label{sec:extension}

We apply ESSENCE unchanged to GLM-4.1V-9B~\cite{glm2025v} (3D-patch) and LLaVA-OneVision-7B~\cite{li2024llavaov} (fixed-size) on TC, VM and LVB (Appendix~E.8).
On GLM-4.1V, ESSENCE improves on downscaling by 4.2 and 3.5 points on VM and LVB and exceeds the full model by 3.4 and 2.5 points at $2.2\times$ faster TTFT, all significantly.
On LLaVA-OneVision, 16 native frames significantly outperform 32 pooled frames on dense videos (1.7 points), but on sparse videos pooling leads by only 0.4 points, not significantly.
ESSENCE-50 remains 1.1 points below the full model (significant) at $2.0\times$ lower time per question: without double exposure to remove, the rule can only choose the less costly reduction.

\section{Conclusion}
\label{sec:conclusion}

We show that on long videos, the accuracy of video LLMs is limited before token selection, by how frames are sampled and tokenized.
We introduce ESSENCE, which allocates token grids to moments with a training-free, tokenizer-specific closed-form rule and releases visual tokens from later LLM layers.
On sparse videos, it stays within 0.7 points of the full model at about $2\times$ faster TTFT, and outperforms five recent accelerators in 58 of 60 comparisons at matched mean TTFT.
We hope these findings encourage future efficient video LLMs to co-design frame sampling and tokenization.

\section*{LLM Usage Statement}
In this work, we used generative AI tools to assist in designing the research methodology and experiments and in implementing the baselines and the evaluation code.
We also used them to edit the manuscript and to improve its readability.
We used Claude (Anthropic) and GPT (OpenAI) for these tasks.
The research questions, the method and all claims were decided by the authors.
We have reviewed all AI-assisted work.
AI-assisted code was executed and tested by the authors, and all reported numbers were checked against the per-question outputs and timing logs.
We take responsibility for the final content of this work, including any content produced with the aid of generative AI.

{
    \small
    \bibliographystyle{ieeenat_fullname}
    \bibliography{main,suppl}
}

\clearpage
\appendix
\crefalias{section}{appendix}
\crefalias{subsection}{appendix}
\crefname{algorithm}{Alg.}{Algs.}
\Crefname{algorithm}{Algorithm}{Algorithms}
\section*{Appendix}
\suppressfloats[t]
\addtocontents{toc}{\protect\setcounter{tocdepth}{2}}
\appendixtoc
\providecommand{\kstar}{k^{*}}
\providecommand{\neff}{n_{\mathrm{eff}}}
\providecommand{\tauc}{\tau_c}
\providecommand{\etat}{\eta_t}
\providecommand{\Wsum}{W_{\Sigma}}
\providecommand{\Wdiff}{W_{\Delta}}

This appendix provides implementation details and additional evidence for the main text.
\Cref{sec:s-impl} details ESSENCE, the evaluation setup, the baselines, and the latency protocol; \cref{sec:s-analysis} expands on the analysis in Sec.~3.
\Cref{sec:s-rule,sec:s-release} examine the allocation rule (Secs.~4.1--4.3) and visual-token release (Sec.~4.4), \cref{sec:s-results} reports per-benchmark results and further experiments for Sec.~5, and \cref{sec:s-limits} discusses limitations.

\noindent\textbf{Notation.}
As in the main text, ESSENCE-$x$ retains $x$\% of the full model's visual tokens: ESSENCE-80, -66, and -42 on Qwen-VL, and ESSENCE-56 and -32 on InternVL3.5.
A video is dense if its median sampling interval satisfies $\Delta t\le\tauc=2$\,s and sparse otherwise; at 32 frames, the four benchmarks comprise 5992 dense and 3425 sparse questions.
Unless stated otherwise, $p$ is two-sided from a video-level sign-flip permutation test with 200\,000 draws, $^*$ denotes $p<0.05$, and no multiple-comparison correction is applied across cells.
Breakdowns by sampling interval, content, or duration are descriptive.

\section{Implementation Details}
\label{sec:s-impl}

\subsection{ESSENCE}
\label{sec:s-essence}

\begin{algorithm}[t]
\caption{ESSENCE for one request.}
\label{alg:s-essence}
\small
\begin{algorithmic}[1]
\Require frames $f_1,\dots,f_T$ with timestamps; budget $B$; tokenizer class and $u$ (Tab.~1); native tokens per grid $b_{\text{native}}$; LLM depth $L$ (layers $0,\dots,L-1$)
\State $\Delta t\gets$ median interval;\quad $\neff\gets T\min(1,\Delta t/\tauc)$ \Comment{Eq.~4}
\State $k_B\gets\lfloor B/b_{\text{native}}\rfloor$
\If{3D-patch}
  \State $k_0\gets\max(T/u,\,k_B)$
  \State $\kstar\gets\mathrm{round}\big(\min(T,\max(\neff,k_0))\big)$ \Comment{Eq.~5}
\ElsIf{$\Delta t>\tauc$} \Comment{Eq.~5 as a threshold, $k_0=k_B$}
  \State $\kstar\gets T$
\Else
  \State $\kstar\gets k_B$
\EndIf
\State $b\gets\min(b_{\text{native}},B/\kstar)$, rounded down to a valid grid
\If{3D-patch}
  \State split the frames into $\kstar$ windows of nearly equal duration
  \State encode two-frame windows as pairs and one-frame windows as stills, resized to $b$ tokens
\Else
  \State encode $\kstar$ evenly spaced frames at the native input
  \State pool each output grid to $b$ tokens if $b<b_{\text{native}}$
\EndIf
\State prefill layers $0,\dots,\ell^*-1$ on all tokens, $\ell^*=\mathrm{round}(2L/3)$
\State remove the visual tokens; run layers $\ell^*,\dots,L-1$ and decoding on text tokens
\end{algorithmic}
\end{algorithm}

\Cref{alg:s-essence} summarizes ESSENCE for a single request.
Every pre-LLM step reads only frame timestamps and the checkpoint attributes in Tab.~1, fixing the encoder input before any frame is encoded.

\noindent\textbf{3D-patch tokenizers.}
Because $\kstar\ge T/u$, each temporal window contains one or two frames.
A single-frame window is duplicated into the patch embedding, which by Eq.~3 encodes that frame alone.
Each input is resized to at most $b\,p^2$ pixels, where $p$ is the pixel side per visual token, reducing the pixel count processed by the encoder.
Each grid retains its frame timestamps.
On Qwen3-VL, the processor formats each grid's timestamp as text (the mean timestamp for a pair, or the single timestamp for a still); on Qwen2.5-VL, we set the seconds per grid to $D/\kstar$ so that M-RoPE temporal positions remain in seconds.
The term $k_B$ allocates budgets exceeding $T/u$ native grids. Because $k_B\le T/u$ at all reported budgets, the main text writes $k_0=T/u$ in Eq.~5.

\noindent\textbf{Fixed-size tokenizers.}
The $\kstar$ frame indices are evenly spaced and include the first and last frames.
When $b<b_{\text{native}}$, the $32\times32$ patch grid of InternVL3.5 is area-averaged to $g\times g$ with $g=2\lfloor\sqrt{b}\rfloor$ before its $2\times2$ pixel shuffle and projector; \cref{sec:s-pool} compares this post-encoder pooling with reducing the input resolution.

\noindent\textbf{Release.}
Layers are indexed from 0, so $\ell^*$ is the index of the first text-only layer and the $\ell^*$ layers before it process all tokens.
At the entry to layer $\ell^*$, we remove the visual-token rows from the hidden states, position embeddings, and attention mask, leaving only text tokens for subsequent layers and decoding; layers before $\ell^*$ retain the visual keys and values in their KV cache for decoding.
Here $\ell^*$ is 24, 19, and 24 on Qwen3-VL, Qwen2.5-VL, and InternVL3.5.
Allocation requires only timestamp arithmetic, and still-frame encoding adds about 10\,ms per request at ESSENCE-66 within the timed region.

\subsection{Evaluation}
\label{sec:s-eval}

\noindent\textbf{Benchmarks.}
TempCompass~\cite{liu2024tempcompass} provides 1580 multiple-choice questions across 410 videos.
MVBench~\cite{li2024mvbench} provides 3800 questions across 19 of its 20 tasks (videos for the fine-grained pose task were unavailable, so all methods are evaluated on the same 3800 questions).
Video-MME~\cite{fu2024videomme} provides 2700 questions across 900 videos evaluated without subtitles, and LongVideoBench~\cite{wu2024longvideobench} provides 1337 validation questions across 753 videos.

\noindent\textbf{Inputs and scoring.}
On each backbone, all methods receive the same 32 uniformly sampled frames (decoded once and shared) and the same prompt listing the options and requesting the option letter.
We score the option letter from the first generated token.
The full model uses its default input: at most $448^2$ pixels per token grid on Qwen-VL, and one 448 tile per frame on InternVL3.5.

\noindent\textbf{Statistics.}
All comparisons are paired on identical questions.
The sign-flip test sums answer-correctness differences per video and randomly flips each video's sum, avoiding treating questions from the same video as independent.
Where noted, we also report exact per-question McNemar tests~\cite{mcnemar1947}.

\subsection{Baselines}
\label{sec:s-baselines}

\begin{table*}[t]
\centering
\scriptsize
\setlength{\tabcolsep}{3pt}
\caption{Code and settings of the baselines on each backbone. Authors' code: the released implementation. Port: our implementation of the released code, or of the paper for ShortV. Indices of layers and blocks start at 0. Released code at commits FastVID a40a109, FlashVID 983cce6, iLLaVA f4d24f8 and EarlyTom 83d80e0.}
\label{tab:s-ports}
\begin{tabular}{@{}p{0.47in}p{0.9in}p{0.9in}p{1.05in}p{3.2in}@{}}
\toprule
Method & Qwen3-VL-8B & Qwen2.5-VL-7B & InternVL3.5-8B & Settings \\
\midrule
FastVID & Port & Port & Port & Retention .27; DySeg $c=8$, $\tau=.84$, ignore .95; STPrune $d=.4$; DTM $p=4$, $\beta=.6$ (authors' Qwen2.5 script) \\
\addlinespace
FlashVID & Authors' functions & Authors' functions & Port calling the authors' functions & Retention .25; segment threshold .9, min 4, complementary; ADTS \texttt{attn\_div} $\alpha=.7$; temporal .8; expansion 1.25. LLM pruning at layer 28, keep .1 (Qwen3-VL, InternVL3.5), or layer 20, keep .3 (Qwen2.5-VL). Authors' scripts for Qwen3-VL and Qwen2.5-VL \\
\addlinespace
iLLaVA & Authors' code and environment & Port of their Qwen3-VL mode 3 & Port of their Qwen3-VL mode 3 & ViT blocks 5--8, keep .9 each; LLM layers 15, 17, 19, 21, keep .85 each; mode 3; recycling ratio .05 (authors' Qwen3-VL script) \\
\addlinespace
EarlyTom & Port & Port & Port & EMA .9, $\tau=.6$, $M=8$, mixing; Stage I at vision blocks 6, 21, 23; $D=0$; retain .25; INNER at layer 18 removes .5 \\
\addlinespace
ShortV & Port & Port & Port & Freeze the 60\% of layers with the lowest contribution, at 448: 1--3, 6, 18--35 on Qwen3-VL (22 of 36); 3--5, 11, 14, 16--27 on Qwen2.5-VL (17 of 28); 2, 3, 5--7, 19--35 on InternVL3.5 (22 of 36) \\
\bottomrule
\end{tabular}
\end{table*}

\Cref{tab:s-ports} summarizes the code and hyperparameters for each baseline.
Because no official release supports InternVL3.5 and several support only one Qwen model, we port the missing variants following the authors' released code wherever available.
Within each backbone, all methods share the same frames, prompt, and input resolution; all scoring, selection, and merging run inside the timed region; and the LLM uses FlashAttention-2~\cite{dao2024flashattention2}.
We use PyTorch 2.8.0 with Transformers 4.57.3 on Qwen-VL and 4.57.1 on InternVL3.5, except for iLLaVA on Qwen3-VL, which requires the authors' modified Transformers 4.57.4 (\cref{sec:s-timing}).

\noindent\textbf{FastVID}~\cite{shao2025fastvid}.
The official code targets Qwen2.5-VL in a modified Transformers 4.51.3.
We preserve its segmentation, selection, and merging with the authors' truncation and tie-breaking, computing saliency via hooks on the native model using their definition (pooled queries, merged keys).
The compressed tokens are fed into the first LLM layer.
On Qwen3-VL, DeepStack~\cite{meng2024deepstack} features are gathered at the selected tokens; on InternVL3.5, scores exclude the CLS token and keys are pooled over the $2\times2$ connector cells.

\noindent\textbf{FlashVID}~\cite{cao2026flashvid}.
On Qwen-VL, we call the authors' functions directly.
Because their dispatcher does not support InternVL3.5, we apply their model-agnostic compression and pruning to the connector output using their Qwen3-VL settings, as both LLMs have 36 Qwen3 layers.

\noindent\textbf{iLLaVA}~\cite{hu2026illava}.
On Qwen3-VL, we run the authors' code in their environment.
On Qwen2.5-VL and InternVL3.5, we port mode 3 of their Qwen3-VL implementation, which selects $2\times2$ merger or pixel-shuffle cells while preserving original positions.
All three backbones use the stage configuration from the authors' Qwen3-VL script rather than the paper's video setting; because layer indices are absolute, they fall relatively deeper in the 28-layer Qwen2.5-VL.

\noindent\textbf{EarlyTom}~\cite{earlytom2026}.
The released code covers only the LLaVA path and an optional Qwen2 LLM, so we port Stage I, Stage II, and INNER to all three backbones using the source defaults and the INNER settings from the authors' script.
The retention ratio of .25 lies within the authors' .10--.25 range.
On Qwen-VL, the temporal unit is the tubelet, and Stage II selects $2\times2$ merger cells with attention recomputed in fp32 because FlashAttention-2 does not return attention weights.

\noindent\textbf{ShortV}~\cite{yuan2025shortv}.
We implement ShortV from the paper.
In frozen layers, visual tokens issue no queries and bypass the FFN, but still supply keys and values to text tokens.
Layer contribution is measured by the KL divergence between the full model and the model with that single layer frozen, evaluated label-free on 16 calibration videos.

\begin{table}[t]
\centering
\scriptsize
\setlength{\tabcolsep}{3.2pt}
\caption{ShortV on InternVL3.5-8B at three freeze ratios with the same layer ranking. Accuracy (\%) and speedups over the full model. The 60\% and ESSENCE-56 rows are those of Tab.~2.}
\label{tab:s-shortv}
\begin{tabular}{@{}lrrrrrrr@{}}
\toprule
Method & TC & VM & MV & LVB & Avg & TTFT & E2E \\
\midrule
Full model & 71.14 & 64.74 & 73.39 & 62.53 & 67.95 & 1.00$\times$ & 1.00$\times$ \\
ShortV, 30\% & 71.27 & 64.41 & 73.50 & 62.30 & 67.87 & 1.19$\times$ & 1.09$\times$ \\
ShortV, 45\% & 71.01 & 64.37 & 72.58 & 61.78 & 67.44 & 1.29$\times$ & 1.14$\times$ \\
ShortV, 60\% & 65.89 & 57.19 & 65.42 & 58.86 & 61.84 & 1.46$\times$ & 1.20$\times$ \\
ESSENCE-56 & 70.63 & 64.33 & 72.58 & 60.73 & 67.07 & 1.95$\times$ & 1.35$\times$ \\
\bottomrule
\end{tabular}
\end{table}

\noindent\textbf{Freeze ratio of ShortV.}
At the authors' 60\% freeze ratio, ShortV drops 6.1 points below the full model on InternVL3.5 (Tab.~2).
Using the same layer ranking, we also evaluate freezing the lowest-contributing 30\% (11 layers: 25--35) and 45\% (16 layers: 2, 20--22, 24--35) of layers (\cref{tab:s-shortv}).
At 30\%, ShortV stays within 0.1 points of the full model and averages 0.80 points above ESSENCE-56, with differences of 0.63, 0.07, 0.92, and 1.57 points on TC, VM, MV, and LVB ($p=0.22$, $0.95$, $0.14$, and $0.057$).
At 45\%, it averages 0.37 points above ESSENCE-56, with differences of 0.38, 0.04, 0.00, and 1.05 points ($p=0.53$, 1.0, 1.0, and $0.23$).
Neither setting is significantly more accurate than ESSENCE-56 on any benchmark, and both are slower (1.19$\times$ and 1.29$\times$ TTFT speedup versus 1.95$\times$ for ESSENCE-56).
Thus, ShortV recovers full-model accuracy on InternVL3.5 at lower freeze ratios and smaller speedups; Tabs.~2 and 3 retain the authors' default.

\noindent\textbf{Validation of the ports.}
With compression disabled, the ports reproduce the full model with real weights (e.g., identical predictions for iLLaVA on Qwen2.5-VL and zero logit difference for FlashVID on InternVL3.5 at retention 1), except FastVID on InternVL3.5 and ShortV on Qwen3-VL and InternVL3.5, which we check with unit tests.
All ports follow the authors' budget rules in token count, and inference is deterministic.
These checks cover structure, budget and determinism, but do not guarantee that a port matches the accuracy of an author-tuned implementation.

\noindent\textbf{LoHi.}
The concurrent LoHi~\cite{dong2026lohi} pairs a dense low-resolution stream with a sparse high-resolution stream.
Because its code was unavailable at submission time, we evaluate its low-resolution stream, which corresponds to our uniform downscaling baseline: all 32 frames downscaled uniformly with ESSENCE's release.
This baseline captures LoHi's core finding that dense low-resolution frames outperform sparse native-resolution frames under a fixed budget.
Uniform downscaling is a strong reference in our experiments (\cref{sec:s-matched}), yet ESSENCE-66 and -56 outperform it on sparse videos across all three models (\cref{tab:s-long}).
Q-Frame and LDDR are detailed alongside their results in \cref{sec:s-qframe}.

\subsection{Latency Measurement}
\label{sec:s-timing}

\noindent\textbf{Requests.}
We measure latency on the same 192 requests across all models and methods: one question per video and 48 per benchmark, sampled with a fixed seed.
The sample is stratified by duration on VM (16 each of short, medium, and long), temporal dimension on TC, task on MV, and duration group on LVB (12 each of 15, 60, 600, and 3600\,s).

\noindent\textbf{Procedure.}
Each model and method runs as a single process on one RTX 5090.
A timing block starts only when no other compute process is active, GPU utilization is at most 10\%, and at least 28\,GiB of memory is free.
Each block runs four warm-up requests followed by the 192 timed requests, executing a TTFT call and a separate E2E call per request.
A second round reverses both method and request order; we compare speeds only within the same model.

\noindent\textbf{Timed region.}
Timing starts from 32 decoded frames in host memory, excluding video decoding and model loading for all methods.
TTFT spans processor and prompt construction, host-to-GPU transfer, vision encoding, any token compression or pruning, LLM prefill, and first-token logits, with GPU synchronization at both ends.
E2E shares the same start and ends after generating exactly 32 greedy tokens with EOS suppressed.

\noindent\textbf{Aggregation.}
Speedup is defined as the full model's mean latency divided by the method's mean latency over the 192 requests, averaged across both rounds so that each benchmark has equal weight.
On Qwen-VL, the median paired per-request speedup ratio lies within 0.1$\times$ of this mean ratio.
On InternVL3.5, the median ratio is higher for ESSENCE (2.18$\times$ and 3.64$\times$ for ESSENCE-56 and -32, versus 1.95$\times$ and 2.72$\times$) because TTFT varies widely across benchmarks on this model.
Because each speedup is computed within its own timing block, its last digit can differ slightly from the ratio of rounded milliseconds in Tab.~2.
Latencies outside Tab.~2 (Tab.~5 and \cref{tab:s-shortv,tab:s-pool,tab:s-qframe}) are measured in separate sessions with the same protocol and requests; across sessions, the speedup of the same ESSENCE configuration varies by at most 0.02$\times$.

\noindent\textbf{iLLaVA on Qwen3-VL.}
On Qwen3-VL, iLLaVA requires the authors' modified Transformers and achieves a 0.65$\times$ TTFT speedup in Tab.~2.
To separate environment overhead from the method itself, we time the authors' environment with both iLLaVA stages disabled on 96 VM requests over two rounds.
Mean TTFT is 411.6\,ms for the full model on the native stack, 594.0\,ms in the authors' environment with iLLaVA disabled, and 656.8\,ms with iLLaVA enabled; mean E2E is 1569.0, 1734.2, and 1796.7\,ms.
The environment alone is 1.44$\times$ slower, and iLLaVA adds 62.8\,ms.
We cannot measure iLLaVA on the native Qwen3-VL stack; our ports on Qwen2.5-VL and InternVL3.5 run on native stacks and reach 1.56$\times$ and 1.04$\times$.
Tab.~3 retains the measured Qwen3-VL speed, pairing iLLaVA with ESSENCE-80.
Even if iLLaVA on the native stack were faster than ESSENCE-80 but not ESSENCE-66 (1.62$\times$ and 1.78$\times$), it would pair with ESSENCE-66, which also outperforms iLLaVA on all four benchmarks by 1.46, 3.22, 0.95, and 0.89 points.

\section{Details of the Analysis in Section 3}
\label{sec:s-analysis}

\subsection{Frame Similarity}
\label{sec:s-similarity}

The similarity metric in Fig.~2a is the phase-correlation peak~\cite{kuglin1975phase} between two frames, defined as the maximum over spatial shifts of the inverse Fourier transform of their normalized cross-power spectrum.
We compute it on $128\times128$ grayscale thumbnails with a Hann window.
Because it relies on spectral phase and searches over translations, it tolerates brightness changes and global shifts.

\subsection{Tokenizer Census}
\label{sec:s-census}

Tab.~1 is compiled directly from public checkpoints without instantiating the models.

\noindent\textbf{Time.}
We read only the patch-embedding tensor of each public checkpoint.
A 5-D tensor of shape $(C,3,2,p,p)$ indicates $u=2$, whereas a 4-D tensor indicates $u=1$.
When $u=2$, $\etat=\|\Wdiff\|/\|\Wsum\|$ across all filters ranges from 0.005 to 0.093, and the cosine similarity between the two temporal kernels ranges from 0.983 to 0.9999.
We classify a tokenizer as 3D-patch if $u>1$ and $\etat<0.1$, a hand-set threshold.

\noindent\textbf{Space.}
The grid size is the spatial side of the vision position embedding excluding any class token; on Qwen3-VL, a $2304\times1152$ tensor yields $48\times48$.
The position grid alone does not determine the class.
A preprocessor configuration specifying \texttt{min\_pixels} and \texttt{max\_pixels}, \texttt{shortest\_edge} and \texttt{longest\_edge}, or \texttt{min\_image\_size} and \texttt{max\_image\_size} declares a variable resolution range.
Otherwise, the declared input size divided by the patch size must match the position grid ($448/14=32$ for InternVL3.5), classifying the tokenizer as fixed-size.

\noindent\textbf{Coverage.}
These criteria classify the eight 3D-patch and eight fixed-size models in Tab.~1.
Aria is flexible along both dimensions and fits neither class.
We exclude Kimi-VL (declares a token limit), Keye-VL (two position tensors), Molmo and Phi-3.5-Vision (configurations do not determine the class), and Gemma 3 (configuration not accessible).
Classifying a checkpoint does not by itself validate the allocation rule on that model.

\subsection{Double Exposure}
\label{sec:s-double}

Sec.~3.3 attributes the accuracy bottleneck on sparse videos to the frame summation performed by 3D patches.
Three experiments on the full model support this mechanism.

\begin{table}[t]
\centering
\scriptsize
\setlength{\tabcolsep}{4pt}
\caption{Pair probes on the full model: accuracy change in points, with $p$ in parentheses.}
\label{tab:s-probes}
\begin{tabular}{@{}llrrr@{}}
\toprule
Model & Probe & Dense (2211) & Sparse (3406) & All (5617) \\
\midrule
Qwen3-VL & Mean & $-1.27$ (0.022) & $-0.29$ (0.56) & $-0.68$ (0.057) \\
 & Swap & $-0.68$ (0.13) & $+0.15$ (0.78) & $-0.18$ (0.60) \\
GLM-4.1V & Mean & $0.00$ (1.00) & $+0.56$ (0.19) & $+0.34$ (0.24) \\
 & Swap & $+0.14$ (0.78) & $+0.32$ (0.46) & $+0.25$ (0.40) \\
\bottomrule
\end{tabular}
\end{table}

\noindent\textbf{Pair probes.}
On TC, VM, and LVB, we either replace each frame pair with its pixel mean or swap the two frames while preserving timestamps (\cref{tab:s-probes}).
Averaging eliminates the difference term in Eq.~2, whereas swapping flips its sign.
On sparse videos, neither perturbation significantly alters accuracy on either model, suggesting limited use of the difference between distant frames.
On dense clips, averaging costs Qwen3-VL 1.27 points ($p=0.022$), consistent with the difference term capturing short-range motion between adjacent frames, which ESSENCE preserves by keeping native pairs on dense videos.

\begin{table}[t]
\centering
\scriptsize
\setlength{\tabcolsep}{4pt}
\caption{Every frame of a sparse video encoded on its own at full size, at twice the tokens: accuracy change against the full model in points, with $p$ in parentheses.}
\label{tab:s-still}
\begin{tabular}{@{}lrrr@{}}
\toprule
Model & VM (2430) & LVB (976) & Both (3406) \\
\midrule
Qwen3-VL & $+2.26$ (0.0008) & $+2.66$ (0.026) & $+2.38$ ($7\times10^{-5}$) \\
Qwen2.5-VL & $+2.55$ (0.0002) & $+3.48$ (0.007) & $+2.82$ ($<10^{-5}$) \\
GLM-4.1V & $+3.25$ ($2\times10^{-5}$) & $+5.23$ ($7\times10^{-5}$) & $+3.82$ ($<10^{-5}$) \\
\bottomrule
\end{tabular}
\end{table}

\noindent\textbf{Every frame on its own.}
Encoding every frame of a sparse video independently as a full-resolution still (doubling the token count) improves accuracy by 2.4 to 3.8 points across all three 3D-patch models (\cref{tab:s-still}).
On Qwen-VL, this variant retains native pairs on dense videos, matching the full model there.
On GLM-4.1V, this variant corresponds to running the allocation rule at twice the budget, which also improves dense videos ($+4.12$ points on 631 questions, $p=0.004$).
This full-resolution still variant exceeds ESSENCE-80, -66, and -42 by 1.32, 1.79, and 3.79 points on Qwen3-VL, and by 1.38, 2.32, and 3.14 points on Qwen2.5-VL, showing that ESSENCE recovers a substantial portion of the frame-separation gain using fewer tokens than the full model.

\noindent\textbf{Number of frames and separation.}
At the token count of ESSENCE-42 on Qwen3-VL, accuracy on the 3406 sparse questions of VM and LVB is 58.72\% with 32 separate frames, 58.04\% with 32 paired frames, 57.43\% with 16 separate frames, and 55.81\% with 16 paired frames.
Both factors contribute significantly: increasing from 16 to 32 frames adds $+1.76$ points ($p=0.0002$) and separating pairs into stills adds $+1.15$ ($p=0.011$), with no significant interaction ($p=0.27$).

\section{Analysis of the Allocation Rule}
\label{sec:s-rule}

\subsection{Statistics of Table 4}
\label{sec:s-coverage}

\begin{table}[t]
\centering
\scriptsize
\setlength{\tabcolsep}{3.5pt}
\caption{The cells of Tab.~4: net answers of the rule over each alternative at the smallest budget, with the questions, videos and $p$.}
\label{tab:s-coverage}
\begin{tabular}{@{}lllrrrr@{}}
\toprule
Model & Case & Alternative & Net & Questions & Videos & $p$ \\
\midrule
Qwen3-VL & Dense & 8 pairs of $413^2$ & $+$43 & 5992 & 4297 & 0.069 \\
 & Dense & 32 stills of $206^2$ & $+$101 & 5361 & 3849 & 0.00002 \\
 & Sparse & 16 pairs of $292^2$ & $+$24 & 3425 & 1221 & 0.27 \\
 & Sparse & 8 pairs of $413^2$ & $+$100 & 3425 & 1221 & 0.00001 \\
\midrule
Qwen2.5-VL & Dense & 8 pairs of $413^2$ & $+$14 & 5992 & 4297 & 0.59 \\
 & Dense & 32 stills of $206^2$ & $+$76 & 5361 & 3849 & 0.0018 \\
 & Sparse & 16 pairs of $292^2$ & $+$74 & 3406 & 1205 & 0.00009 \\
 & Sparse & 8 pairs of $413^2$ & $+$102 & 3425 & 1221 & 0.00001 \\
\midrule
InternVL3.5 & Dense & 32 frames, pooled & $+$84 & 5361 & 3849 & 0.00033 \\
 & Dense & 32 frames of 252 & $+$143 & 5992 & 4297 & 0.00001 \\
 & Sparse & 10 frames of 448 & $+$82 & 3406 & 1205 & 0.00037 \\
 & Sparse & 32 frames of 252 & $+$42 & 3425 & 1221 & 0.037 \\
\bottomrule
\end{tabular}
\end{table}

Each cell in Tab.~4 compares the allocation rule against an alternative on the shared subset of questions (\cref{tab:s-coverage}).
The rule cells are the ESSENCE-42 and ESSENCE-32 runs. On dense 3D-patch videos, the rule encodes 16 pairs on 5441 of the 5992 dense questions and 17 to 32 grids, mixing pairs and stills, on the other 551, whose intervals lie between 1 and 2\,s; Tab.~4 labels the row by the first case.
Two dense-video alternatives (32 stills and 32 pooled frames) were evaluated on TC and MV only, and two sparse-video alternatives (Qwen2.5-VL with 16 pairs and InternVL3.5 with 10 native frames) on VM and LVB only.

\noindent\textbf{Token counts.}
All alternatives target the same budget as the rule, with minor token-count differences from rounding grids to whole patches.
On Qwen3-VL, the rule averages 1285 and 1336 visual tokens per dense and sparse question; 8 pairs of $413^2$ use 1221 and 1296, 32 stills of $206^2$ use 1322 on dense videos, and 16 pairs of $292^2$ use 1303 on sparse videos (95\% to 103\% of the rule).
On Qwen2.5-VL, alternative prompts are 1\% to 15\% longer than the rule's (1575 and 1556 tokens); the largest gap, 8 pairs of $413^2$ on sparse videos, favors the alternative.
On InternVL3.5, 10 native frames produce 2560 visual tokens while both 32-frame alternatives produce 2592 and add roughly 170 text tokens for frame labels.

\noindent\textbf{Interaction.}
Sec.~5.4 shows that the optimal dense-video policy reverses across tokenizer classes.
For each dense video, we compute the net correct answers gained by lowering spatial resolution rather than dropping frames on the 3D-patch models (the rule minus 8 pairs of $413^2$) and on InternVL3.5 (32 frames of 252 minus 10 frames of 448), then test the difference between classes by sign-flipping.
Across 4297 videos, the interaction totals 186 answers between Qwen3-VL and InternVL3.5 ($p\approx10^{-5}$) and 157 between Qwen2.5-VL and InternVL3.5 ($p=6\times10^{-5}$).
Replacing 32 frames of 252 with pooled frames across the 3849 videos of TC and MV yields differences of 110 ($p=0.0004$) and 86 answers ($p=0.009$).

\subsection{Moments from Timestamps or from Content}
\label{sec:s-content}

Eq.~4 estimates the number of moments from timestamps alone.
Because a content-based estimator could merge static scenes regardless of duration and avoid pairing across shot cuts, we compare our timestamp rule against two content-based variants across all 9417 questions.

\noindent\textbf{Per-pair test.}
Adjacent frames are paired only if their phase-correlation peak exceeds the 99th percentile of the unrelated-frame null distribution ($p<0.01$).
Relative to the timestamp rule on Qwen3-VL, this changes correct answers on TC, VM, MV, and LVB by $+6$, $-18$, $+9$, and $+10$ at ESSENCE-66, and by $+4$, $+12$, $-8$, and $-2$ at ESSENCE-42.
On InternVL3.5, where coherent pairs are pixel-averaged into a single frame, LVB accuracy drops by 0.75 and 1.05 points at ESSENCE-56 and -32.

\noindent\textbf{Content-based $\neff$.}
Estimating $\neff$ from inter-frame correlation rather than timestamps changes correct answers on the four benchmarks by $-17$, 0, $+6$, and $-7$ at ESSENCE-66, and by $-24$, $+1$, $+9$, and $+4$ at ESSENCE-42 on Qwen3-VL.
On InternVL3.5, the changes are $-6$, $-9$, $+8$, and 0 at ESSENCE-56, and $-9$, $-9$, $-29$, and $+1$ at ESSENCE-32, while also running slower at ESSENCE-56.

Neither content-based variant gains more than 7 answers out of 9417 in total on any model, and both require inspecting frame pixels before allocation, whereas timestamps do not.

\begin{table}[t]
\centering
\scriptsize
\setlength{\tabcolsep}{2.4pt}
\caption{Static sparse videos and on-screen text on Qwen3-VL. Accuracy (\%) of ESSENCE-66, uniform downscaling at the same budget and the full model, and the differences of ESSENCE in points.}
\label{tab:s-static}
\begin{tabular}{@{}lrrrrrr@{}}
\toprule
Subset & $n$ & Ours & Down. & Full & $-$Down. & $-$Full \\
\midrule
VM sparse, static & 315 & 67.62 & 66.67 & 68.57 & $+0.95$ & $-0.95$ \\
VM sparse, dynamic & 2115 & 62.13 & 60.19 & 61.32 & $+1.94^*$ & $+0.80$ \\
LVB sparse, static & 200 & 53.50 & 50.00 & 53.50 & $+3.50$ & 0.00 \\
LVB sparse, dynamic & 776 & 55.93 & 54.25 & 55.15 & $+1.68$ & $+0.77$ \\
VM OCR questions & 139 & 68.35 & 62.59 & 67.63 & $+5.76$ & $+0.72$ \\
\bottomrule
\end{tabular}
\end{table}

\noindent\textbf{Static videos and on-screen text.}
On static sparse videos, the timestamp rule encodes lower-resolution stills just as on dynamic videos, even though paired frames would remain coherent.
Defining a sparse video as static when the per-pair test marks at least 8 of its 16 pairs as coherent, we compare ESSENCE-66 against uniform downscaling at the same budget, which retains native pairs (\cref{tab:s-static}).
On the static sparse subsets, ESSENCE is 0.95 points below the full model on VM and tied with it on LVB; its differences from uniform downscaling are not significant ($p=0.68$ and 0.21).
On the 139 OCR questions of VM, ESSENCE exceeds downscaling by 5.76 points ($p=0.13$) and the full model by 0.72 points.
Thus, in our benchmarks, encoding stills shows no significant penalty against downscaling on static scenes or on-screen text.

\subsection{Sensitivity to \texorpdfstring{$\tauc$}{tau\_c}}
\label{sec:s-tau}

\begin{table}[t]
\centering
\scriptsize
\setlength{\tabcolsep}{3pt}
\caption{Sensitivity of the rule: correct answers of ESSENCE in the main table, and the change from halving or doubling $\tauc$ or from the smooth fixed-size rule. In parentheses: questions whose input changes.}
\label{tab:s-robust}
\begin{tabular}{@{}lrrrr@{}}
\toprule
 & TC & VM & MV & LVB \\
\midrule
\multicolumn{5}{@{}l}{\textit{Qwen3-VL, ESSENCE-66}} \\
Main table ($\tauc=2$\,s) & 1172 & 1741 & 2584 & 812 \\
$\tauc=1$\,s & $-2$ (349) & $+1$ (255) & $-4$ (1414) & $-2$ (155) \\
$\tauc=4$\,s & 0 (16) & $-9$ (843) & $-7$ (241) & $-5$ (97) \\
\midrule
\multicolumn{5}{@{}l}{\textit{InternVL3.5, ESSENCE-56}} \\
Main table ($\tauc=2$\,s) & 1116 & 1737 & 2758 & 812 \\
$\tauc=1$\,s & 0 (16) & 0 (216) & $+4$ (250) & $+4$ (98) \\
$\tauc=4$\,s & 0 (0) & $-17^*$ (627) & 0 (19) & 0 (0) \\
Smooth rule & 0 (0) & $-1$ (204) & $-1$ (55) & $+3$ (84) \\
\midrule
\multicolumn{5}{@{}l}{\textit{InternVL3.5, ESSENCE-32}} \\
Main table & 1113 & 1687 & 2757 & 804 \\
Smooth rule & $+2$ (212) & $+1$ (252) & $-3$ (1177) & $+3$ (138) \\
\bottomrule
\end{tabular}
\end{table}

The constant $\tauc=2$\,s is derived from frame decorrelation in Sec.~3.2 and shared across all models.
\Cref{tab:s-robust} halves and doubles it for ESSENCE-66 on Qwen3-VL (altering $\neff$) and ESSENCE-56 on InternVL3.5 (shifting the threshold between $k_B$ native frames and $T$ pooled frames).
Only questions whose model input changes are re-evaluated, since inference is deterministic (\cref{sec:s-baselines}).

On Qwen3-VL, no change is statistically significant or exceeds 9 answers (0.33 points), and $\tauc=2$\,s is best or within one answer of the best on every benchmark.
On InternVL3.5, $\tauc=1$\,s pools all 32 frames for questions with $1<\Delta t\le2$\,s, gaining 8 answers overall ($p=0.45$ on MV and 0.12 on LVB) at 1.4$\times$ the latency per affected question.
Setting $\tauc=4$\,s retains 18 of 32 frames on the 627 VM questions with $2<\Delta t\le4$\,s, losing 17 answers ($-0.63$ points, $p=0.045$) while reducing latency on those questions by 28\% (0.60 versus 0.83\,s).
Halving or doubling $\tauc$ thus shifts accuracy by at most 0.63 points, with the only significant change favoring $\tauc=2$\,s.

\subsection{Continuous Rule and Step}
\label{sec:s-smooth}

\noindent\textbf{3D-patch tokenizers.}
At $T=32$, a binary gate that switches directly from $T/u$ native pairs for $\Delta t\le\tauc$ to $T$ stills otherwise differs from Eq.~5 only when $1<\Delta t\le2$\,s, where $T/u<\neff<T$.
On Qwen3-VL at ESSENCE-66, the continuous rule yields 1172, 1741, 2584, and 812 correct answers versus 1172, 1741, 2578, and 807 for the binary gate.
The continuous rule leads by 6 answers on MV and 5 on LVB ($p=0.27$), entirely from questions in this transition band.
While separating sparse moments provides the primary gain, the continuous rule smoothly interpolates between the two regimes.

\noindent\textbf{Fixed-size tokenizers.}
On fixed-size tokenizers, we apply Eq.~5 as a step threshold from $k_B$ native frames to $T$ pooled frames at $\Delta t=\tauc$.
Applied literally, Eq.~5 yields a smooth variant that retains $\mathrm{round}(\neff)$ frames whenever $k_B<\neff\le T$, encodes them at 448, and pools them to fit the budget, gradually increasing the frame count from $k_B$ at $\Delta t=\tauc k_B/T$ to $T$ at $\tauc$.
This smooth variant changes correct answers across the four benchmarks by 0, $-1$, $-1$, and $+3$ at ESSENCE-56 and by $+2$, $+1$, $-3$, and $+3$ at ESSENCE-32 (none significant; \cref{tab:s-robust}), while increasing latency by 1.05--1.64$\times$ per affected question because the encoder processes more frames at 448.
Because frames spaced closer than $\tauc$ are largely redundant and native frames outperform pooled frames on dense videos at equal budget (Tab.~4), the step threshold saves encoding time without significant accuracy loss.

\subsection{Pooling or Smaller Input on InternVL3.5}
\label{sec:s-pool}

\begin{table}[t]
\centering
\scriptsize
\setlength{\tabcolsep}{3pt}
\caption{Pooling after the encoder against a smaller input on InternVL3.5. Correct answers of 1580, 2700, 3800 and 1337; Avg: mean accuracy (\%); speedups over the full model. TC inputs are identical within a budget. ESSENCE-56 and -32 are the pooling rows.}
\label{tab:s-pool}
\begin{tabular}{@{}llrrrrrrr@{}}
\toprule
Budget & Variant & TC & VM & MV & LVB & Avg & TTFT & E2E \\
\midrule
66\% & Smaller input & 1120 & 1712 & 2778 & 807 & 66.94 & 1.83$\times$ & 1.32$\times$ \\
 & Pooling & 1120 & 1694 & 2777 & 805 & 66.73 & 1.71$\times$ & 1.29$\times$ \\
\midrule
56\% & Smaller input & 1116 & 1719 & 2760 & 805 & 66.79 & 2.10$\times$ & 1.39$\times$ \\
 & Pooling & 1116 & 1737 & 2758 & 812 & 67.07 & 1.95$\times$ & 1.35$\times$ \\
\midrule
32\% & Smaller input & 1113 & 1667 & 2755 & 784 & 65.83 & 3.36$\times$ & 1.61$\times$ \\
 & Pooling & 1113 & 1687 & 2757 & 804 & 66.40 & 2.72$\times$ & 1.52$\times$ \\
\bottomrule
\end{tabular}
\end{table}

On sparse videos, the fixed-size rule retains all 32 frames with fewer tokens per frame.
This reduction can be performed before the encoder by resizing inputs to 364, 336, or 252 (interpolating the learned position grid), or after the encoder by encoding at 448 and area-averaging the patch grid (Sec.~4.3).
Both options produce 169, 144, and 81 tokens per frame at the 66\%, 56\%, and 32\% budgets, and both retain $\lfloor B/256\rfloor$ native frames on dense videos (\cref{tab:s-pool}).

Resizing the input is faster at every budget because the encoder processes fewer pixels (e.g., 1.95$\times$ versus 1.65$\times$ on VM at 56\%, and 2.88$\times$ versus 1.94$\times$ at 32\%).
However, post-encoder pooling yields more correct answers on VM and LVB at the 56\% and 32\% budgets: $+25$ ($p=0.14$) and $+40$ ($p=0.046$) in total, with exact McNemar $p=0.14$ and 0.041.
Moreover, ESSENCE-56 with pooling is both faster and more accurate than either 66\% variant.
Our conclusions against the baselines hold under either choice: with smaller inputs, the speed-matched budget is the 66\% variant for all five baselines, where ESSENCE leads in 57 of 60 cells in Tab.~3, ties in 1, and trails in 2 (on LVB, by 0.37 and 0.75 points against FlashVID and iLLaVA), with 44 significant wins and no significant losses, compared to 58 wins, 1 tie, 1 loss, and 46 significant wins in Tab.~3.

\section{Analysis of Release}
\label{sec:s-release}

\subsection{Calibration and a Stricter Criterion}
\label{sec:s-horizon}

\noindent\textbf{Calibration questions.}
The curve $f(\ell)$ in Fig.~5 is measured on 192 unlabeled calibration questions: 64 videos disjoint from all evaluation sets, each paired with three questions from six generic templates.
At each release layer, we count questions whose top-1 option letter differs from the full model's prediction.
A control switches only the LLM attention kernel from FlashAttention-2 to mathematically equivalent SDPA, isolating flips caused by numerical precision.

\begin{table}[t]
\centering
\scriptsize
\setlength{\tabcolsep}{2.6pt}
\caption{Flips of the 192 calibration questions at $\ell^*$ with one-sided 95\% Clopper--Pearson bounds over questions and over videos, flips of the control, the depth $\ell_0$ of the null-calibrated criterion, and the same criterion on full-vocabulary KL. Layers are indexed from 0.}
\label{tab:s-horizon}
\begin{tabular}{@{}lrrrrrrrr@{}}
\toprule
Model & $L$ & $\ell^*$ & Flips & Bounds (\%) & Control & $q$ & $\ell_0$ ($\ell_0/L$) & KL \\
\midrule
Qwen3-VL & 36 & 24 & 1 (0.52\%) & 2.45 / 7.20 & 1 & 3 & 22 (0.61) & 28 \\
Qwen2.5-VL & 28 & 19 & 4 (2.08\%) & 4.70 / 13.73 & 0 & 0 & 23 (0.82) & 21 \\
InternVL3.5 & 36 & 24 & 2 (1.04\%) & 3.24 / 9.51 & 1 & 3 & 33 (0.92) & 35 \\
\bottomrule
\end{tabular}
\end{table}

\noindent\textbf{Bounds on the flip rate.}
Because the three questions per video are not independent, we bound the flip rate at $\ell^*$ in two ways (\cref{tab:s-horizon}).
Across all 192 questions, one-sided 95\% Clopper--Pearson bounds~\cite{clopper1934} are 2.45\%, 4.70\% and 3.24\%.
Across the 64 videos, counting a video if any of its questions flips bounds the per-question rate from above at 7.20\%, 13.73\% and 9.51\%.
Every flip at $\ell^*$ occurs in a distinct video, so the wider video-level bounds reflect only the smaller sample size.
Sec.~5.5 therefore reports the benchmark flip rate (\cref{sec:s-flips}), which bounds the accuracy change without labels, and the calibration questions serve only to select $\ell^*$.

\noindent\textbf{A null-calibrated criterion.}
A per-model alternative releases tokens at the shallowest layer where flips are indistinguishable from numerical noise.
Let $q$ be the 95th percentile of 10\,000 bootstrap means of the control's flips; this criterion selects the shallowest measured layer $\ell_0$ where flips at $\ell_0$ and all deeper measured layers are at most $q$.
\Cref{tab:s-horizon} compares $\ell_0$ with $\ell^*=\mathrm{round}(2L/3)$: it is shallower on Qwen3-VL and deeper on the other two models.
With 192 questions, a single flip is 0.52\% and can determine $\ell_0$.
On Qwen2.5-VL, the control has zero flips ($q=0$), so a single flip at layer 22 sets $\ell_0=23$; InternVL3.5 enters the null band at layer 24 but exits at layer 32 with 4 flips against $q=3$.
We retain a single relative ratio across all models to avoid per-model search; at $\ell^*$, the calibration questions flip 1, 4 and 2 times.

\subsection{Answer Changes on the Benchmarks}
\label{sec:s-flips}

Across the 9417 benchmark questions, releasing tokens in ESSENCE-66 or -56 changes the predicted option letter on 0.97\% of questions on Qwen3-VL, 3.72\% on Qwen2.5-VL and 1.67\% on InternVL3.5, with 95\% video-bootstrap intervals of 0.77--1.17, 3.34--4.11 and 1.40--1.93.
Because the first token decides the answer, these label-free rates bound the accuracy change from release on the benchmarks.
All three lie within the video-level calibration bounds, though Qwen2.5-VL exceeds its 2.08\% calibration estimate; the calibration questions thus set the release depth rather than predict exact flip rates on new distributions.
In accuracy, these flips largely cancel: release alters no overall total in Tab.~5 significantly, and changes only 2 of the 12 benchmark cells in \cref{tab:s-ablation}, in opposite directions.

\subsection{Reductions before \texorpdfstring{$\ell^*$}{l*}}
\label{sec:s-before}

ESSENCE leaves the layers before $\ell^*$ intact and reduces tokens before the vision encoder.
On Qwen3-VL, none of the reductions we evaluated inside these layers (skipping the visual MLP, reusing or predicting visual states, or letting visual tokens only provide keys and values) outperforms lowering the input resolution.
Each layer changes about one-third of the visual hidden state (relative update 0.26--0.39), a change poorly predicted by neighboring tokens, earlier layers or adjacent tubelets, and skipping or freezing operations saves little latency for the first-token KL divergence it adds.

\subsection{Attention after Release and Task Scope}
\label{sec:s-scope}

\noindent\textbf{Attention after release.}
In the full model, the answer token still allocates 3--10\% of its attention to visual tokens in the late layers of Qwen-VL, and 20--36\% in layers 23--34 of InternVL3.5, where visual tokens contribute 0.2--0.34 of the attention output.
Late layers thus continue to attend to visual tokens, and release relies on the observation that this attention rarely alters the final decision, as measured directly in Fig.~5.

\noindent\textbf{Task scope.}
The bound in Sec.~4.4 applies when the first generated token determines the answer, as in our four multiple-choice benchmarks.
In free-form generation, subsequent tokens may still need visual features, and because released tokens are omitted from the KV cache from layer $\ell^*$ onward, follow-up turns in multi-turn dialogue can attend to them only in earlier layers.

\begin{table}[t]
\centering
\scriptsize
\setlength{\tabcolsep}{3pt}
\caption{Release on TempCompass caption generation (2004 captions of up to 256 tokens, greedy). Accuracy (\%) of a text-only judge that answers the multiple-choice question of each caption from the caption alone, mean caption length in words, and the net answers and $p$ of release.}
\label{tab:s-caption}
\begin{tabular}{@{}llrrrrr@{}}
\toprule
 & & \multicolumn{2}{c}{Accuracy} & \multicolumn{2}{c}{Words} & \\
\cmidrule(lr){3-4}\cmidrule(lr){5-6}
Model & Input & w/o & with & w/o & with & Net ($p$) \\
\midrule
Qwen3-VL & 16 pairs, $448^2$ & 67.66 & 68.46 & 36.6 & 27.6 & $+16$ (0.39) \\
 & 16 pairs, $365^2$ & 67.96 & 68.86 & 36.6 & 27.5 & $+18$ (0.26) \\
 & 16 pairs, $292^2$ & 67.37 & 69.16 & 34.5 & 26.6 & $+36$ (0.020) \\
InternVL3.5 & 32 frames, 336 & 60.73 & 64.67 & 9.2 & 8.1 & $+79$ ($<10^{-3}$) \\
 & 32 frames, 252 & 63.12 & 64.87 & 9.0 & 8.0 & $+35$ (0.044) \\
\bottomrule
\end{tabular}
\end{table}

\noindent\textbf{Caption generation.}
To evaluate release beyond the first token, we test on the TempCompass caption-generation task: the model captions each of the 410 videos along a target temporal dimension, and a text-only judge (Qwen2.5-VL-7B) answers the corresponding multiple-choice question from the generated caption alone, following the official protocol.
\Cref{tab:s-caption} compares matched inputs with and without layer-24 release on Qwen3-VL and InternVL3.5.
Release alters most captions (only 15--44\% remain identical) and shortens Qwen3-VL captions by about 25\%, yet judged accuracy never drops across the five settings and increases significantly in three.
The judge evaluates whether a caption captures the queried temporal event rather than its length or descriptive detail.
For long-form generation or multi-turn dialogue where fine detail is needed, release can be disabled: budget allocation and native encoding depend only on timestamps, and without release ESSENCE-66 and -56 still achieve 1.44, 1.59 and 1.64$\times$ TTFT speedups with no significant accuracy change (Tab.~5).

\section{Additional Results}
\label{sec:s-results}

\subsection{Per-Benchmark Ablation}
\label{sec:s-ablation}

\begin{table}[t]
\centering
\scriptsize
\setlength{\tabcolsep}{4pt}
\caption{Net correct answers of ESSENCE, at the budget in the second column, over each variant.}
\label{tab:s-ablation}
\begin{tabular}{@{}llrrrrr@{}}
\toprule
Variant & Budget & TC & VM & MV & LVB & All \\
\midrule
\multicolumn{7}{@{}l}{\textit{Qwen3-VL-8B}} \\
No release & 66\% & +4 & +6 & +3 & +1 & +14 \\
No allocation & 66\% & 0 & +44$^{*}$ & +11$^{*}$ & +25$^{*}$ & +80$^{*}$ \\
Uniform frames & 42\% & +23 & +72$^{*}$ & $-$5 & +33$^{*}$ & +123$^{*}$ \\
VTW only & 66\% & $-$5 & +14 & $-$30$^{*}$ & +12 & $-$9 \\
\midrule
\multicolumn{7}{@{}l}{\textit{Qwen2.5-VL-7B}} \\
No release & 66\% & $-$11 & $-$7 & $-$4 & +14$^{*}$ & $-$8 \\
No allocation & 66\% & 0 & +46$^{*}$ & +2 & +14 & +62$^{*}$ \\
Uniform frames & 42\% & $-$22 & +113$^{*}$ & 0 & +5 & +96$^{*}$ \\
VTW only & 66\% & $-$22$^{*}$ & +12 & $-$51$^{*}$ & +1 & $-$60$^{*}$ \\
\midrule
\multicolumn{7}{@{}l}{\textit{InternVL3.5-8B}} \\
No release & 56\% & $-$4 & +12 & $-$16$^{*}$ & 0 & $-$8 \\
No allocation & 56\% & +8 & +26 & +40$^{*}$ & +6 & +80$^{*}$ \\
Uniform frames & 32\% & $-$3 & 0 & $-$1 & +5 & +1 \\
VTW only & 56\% & $-$9 & $-$1 & $-$41$^{*}$ & $-$19 & $-$70$^{*}$ \\
\bottomrule
\end{tabular}
\end{table}

\Cref{tab:s-ablation} breaks down the ablation in Tab.~5 by benchmark and adds two baselines.
Uniform frames samples 16 frames on Qwen-VL and 18 on InternVL3.5 at native resolution, paired with the nearest faster ESSENCE budgets (-42 and -32).
VTW only retains all full-model tokens and releases them at the layer selected by the VTW criterion~\cite{lin2025boosting}, where mean first-token KL on calibration videos drops below 0.003 (layers 24, 20 and 33, versus 24, 19 and 24 for ESSENCE).

Removing allocation costs 62 to 80 correct answers per model, mostly on VM and LVB on Qwen-VL and half of them on MV on InternVL3.5.
On TC, the net difference is 0 on Qwen-VL because both variants receive identical inputs when $\Delta t\le1$\,s.
Removing release changes only two cells significantly, in opposite directions: Qwen2.5-VL on LVB ($+14$) and InternVL3.5 on MV ($-16$).
Uniform frames trails ESSENCE by 123 and 96 answers on Qwen-VL and matches it on InternVL3.5, where dropping frames is the preferred dense-video reduction (Tab.~4).
Because VTW only keeps all visual tokens before release, it leads on MV across all three models and overall on Qwen2.5-VL and InternVL3.5 by 60 and 70 answers.
However, it is substantially slower: its TTFT is 316.8 versus 224.8\,ms for ESSENCE-66 on Qwen3-VL, and 1269.1 versus 684.9\,ms for ESSENCE-56 on InternVL3.5 (\cref{tab:s-stage}).

\subsection{Per-Benchmark Comparison at Matched Speed}
\label{sec:s-matched}

\begin{table}[t]
\centering
\scriptsize
\setlength{\tabcolsep}{4pt}
\caption{Gain of ESSENCE over each baseline in points, at the largest budget at least as fast on mean TTFT (Tab.~3). $^*$: $p<0.05$ with 100\,000 draws. $^\dagger$: ESSENCE is slower than the baseline on this benchmark.}
\label{tab:s-matched}
\begin{tabular}{@{}llrrrr@{}}
\toprule
Baseline & Budget & TC & VM & MV & LVB \\
\midrule
\multicolumn{6}{@{}l}{\textit{Qwen3-VL-8B}} \\
FastVID & 66\% & +4.24$^{*}$ & +5.56$^{*}$ & +2.03$^{*}$ & +3.22$^{*}$ \\
FlashVID & 80\% & +2.28$^{*}$ & +4.00$^{*}$ & +2.03$^{*\dagger}$ & +2.69$^{*}$ \\
iLLaVA & 80\% & +1.46 & +3.48$^{*}$ & +1.71$^{*}$ & +1.72 \\
EarlyTom & 66\% & +2.41$^{*}$ & +3.63$^{*}$ & +1.76$^{*\dagger}$ & +4.86$^{*}$ \\
ShortV & 80\% & +1.58 & +4.07$^{*}$ & +1.21$^{*\dagger}$ & +0.30 \\
\midrule
\multicolumn{6}{@{}l}{\textit{Qwen2.5-VL-7B}} \\
FastVID & 80\% & +2.72$^{*\dagger}$ & +4.93$^{*}$ & +1.61$^{*\dagger}$ & +2.69$^{*}$ \\
FlashVID & 80\% & +2.41$^{*}$ & +3.22$^{*}$ & +0.13$^{\dagger}$ & +2.17$^{*}$ \\
iLLaVA & 80\% & +0.70 & +2.89$^{*}$ & +0.47$^{\dagger}$ & +2.92$^{*}$ \\
EarlyTom & 66\% & +1.46 & +3.48$^{*}$ & +1.95$^{*\dagger}$ & +1.65 \\
ShortV & 80\% & +3.92$^{*}$ & +5.00$^{*}$ & +2.66$^{*}$ & +2.99$^{*}$ \\
\midrule
\multicolumn{6}{@{}l}{\textit{InternVL3.5-8B}} \\
FastVID & 56\% & +2.09$^{*}$ & +2.70$^{*\dagger}$ & +4.76$^{*}$ & +0.37 \\
FlashVID & 56\% & +2.03$^{*}$ & +1.93$^{*}$ & +5.82$^{*}$ & 0.00 \\
iLLaVA & 56\% & +0.19 & +1.22$^{*}$ & +1.42$^{*}$ & $-$0.37 \\
EarlyTom & 56\% & +1.33$^{*}$ & +3.52$^{*\dagger}$ & +3.66$^{*}$ & +2.62$^{*}$ \\
ShortV & 56\% & +4.75$^{*}$ & +7.15$^{*}$ & +7.16$^{*}$ & +1.87 \\
\bottomrule
\end{tabular}
\end{table}

\Cref{tab:s-matched} lists the 60 comparisons of Tab.~3 along with the matched ESSENCE budgets.
Clustering questions by video (410, 900, 3455 and 753 videos on TC, VM, MV and LVB), 10, 15, 13 and 8 of the 15 comparisons per benchmark are significant (46 in total), and 27 survive Bonferroni correction at $0.05/60$.
Exact per-question McNemar tests yield 48 significant wins, 31 of them below the Bonferroni level.
The only two non-winning cells occur on InternVL3.5 on LVB: a tie with FlashVID and a $-0.37$-point difference against iLLaVA ($p=0.72$).

\noindent\textbf{Speed per benchmark.}
Budgets are matched by mean TTFT across the four benchmarks.
On individual benchmarks, ESSENCE is slower than the paired baseline in the 10 cells marked $^\dagger$ by 0.3 to 10.6\% (7 on MV, 2 on VM and 1 on TC); the largest gap is ESSENCE-66 versus EarlyTom on MV with Qwen2.5-VL (262.9 vs.\ 237.7\,ms per request).
Eight of these 10 cells are significant, leaving 38 significant wins where ESSENCE is also at least as fast on that specific benchmark.

\noindent\textbf{Uniform downscaling.}
Combined with release at the ESSENCE-66 budget (a 336 tile at 56\% on InternVL3.5), uniform downscaling outperforms the five baselines in 17, 18 and 16 of the 20 comparisons on Qwen3-VL, Qwen2.5-VL and InternVL3.5 (11, 7 and 11 significant), with 1 tie and 8 losses.
On VM, it leads in all 15 comparisons (11 significant), compared with 15 significant leads for ESSENCE.

\subsection{Comparison with Q-Frame and LDDR}
\label{sec:s-qframe}

\begin{table*}[t]
\centering
\scriptsize
\setlength{\tabcolsep}{5pt}
\caption{Q-Frame and LDDR on Qwen3-VL-8B and Qwen2.5-VL-7B. Accuracy (\%), mean visual tokens per request, and speedups over the full model on the 192 timed requests. The ESSENCE rows are those of Tab.~2.}
\label{tab:s-qframe}
\begin{tabular}{@{}lrrrrrrrrrrr@{}}
\toprule
 & \multicolumn{5}{c}{Accuracy} & \multicolumn{4}{c}{Visual tokens} & & \\
\cmidrule(lr){2-6}\cmidrule(lr){7-10}
Method & TC & VM & MV & LVB & Avg & TC & VM & MV & LVB & TTFT & E2E \\
\midrule
\multicolumn{12}{@{}l}{\textit{Qwen3-VL-8B}} \\
Full model & 74.49 & 64.04 & 68.76 & 59.91 & 66.80 & 3007 & 2879 & 2453 & 2883 & 1.00$\times$ & 1.00$\times$ \\
Q-Frame, official & 63.92 & 61.78 & 58.63 & 58.79 & 60.78 & 496 & 1626 & 1131 & 1754 & 0.54$\times$ & 0.83$\times$ \\
Q-Frame, fp32 ranking & 63.67 & 61.44 & 58.82 & 59.09 & 60.76 & 496 & 1626 & 1131 & 1754 & -- & -- \\
Q-Frame, resolution-matched & 65.38 & 62.11 & 62.13 & 59.54 & 62.29 & 1531 & 1530 & 1266 & 1531 & 0.55$\times$ & 0.83$\times$ \\
\quad excl.\ selection time & & & & & & & & & & 1.48$\times$ & 1.09$\times$ \\
\quad 128 candidates & -- & 61.07 & -- & 59.16 & -- & -- & 1598 & -- & 1599 & -- & -- \\
LDDR & 69.05 & 64.22 & 63.89 & 60.73 & 64.48 & 1901 & 1880 & 1890 & 1827 & 0.52$\times$ & 0.82$\times$ \\
\quad 128 candidates & -- & 65.48 & -- & 62.98 & -- & -- & 2005 & -- & 1978 & -- & -- \\
ESSENCE-80 & 74.18 & 64.74 & 68.76 & 61.56 & 67.31 & & & & & 1.62$\times$ & 1.11$\times$ \\
ESSENCE-66 & 74.18 & 64.48 & 68.00 & 60.73 & 66.85 & & & & & 1.78$\times$ & 1.13$\times$ \\
ESSENCE-42 & 73.80 & 62.26 & 66.89 & 58.71 & 65.42 & & & & & 2.32$\times$ & 1.18$\times$ \\
\midrule
\multicolumn{12}{@{}l}{\textit{Qwen2.5-VL-7B}} \\
Full model & 71.84 & 59.67 & 64.26 & 55.65 & 62.85 & 3906 & 4026 & 3264 & 4027 & 1.00$\times$ & 1.00$\times$ \\
Q-Frame, resolution-matched & 68.04 & 57.74 & 59.11 & 57.37 & 60.56 & 2031 & 2027 & 1656 & 2028 & 0.63$\times$ & 0.82$\times$ \\
LDDR & 69.87 & 59.04 & 60.87 & 55.87 & 61.41 & 2508 & 2484 & 2490 & 2414 & 0.61$\times$ & 0.81$\times$ \\
ESSENCE-80 & 71.52 & 60.78 & 62.66 & 56.77 & 62.93 & & & & & 1.59$\times$ & 1.17$\times$ \\
ESSENCE-66 & 70.38 & 60.00 & 63.11 & 55.42 & 62.23 & & & & & 1.90$\times$ & 1.23$\times$ \\
ESSENCE-42 & 69.49 & 59.30 & 62.03 & 54.82 & 61.41 & & & & & 2.84$\times$ & 1.33$\times$ \\
\bottomrule
\end{tabular}
\end{table*}

Like ESSENCE, Q-Frame~\cite{zhang2025qframe} acts before vision encoding.
It scores candidate frames against the question using Long-CLIP~\cite{zhang2024longclip}, ranks them, and encodes the top frames at three spatial resolutions.
We port the official Qwen2-VL code (Long-CLIP-L at 336\,px) to Qwen3-VL and Qwen2.5-VL using its 32-candidate defaults.
The question stem (without options) is scored against all 32 frames, ranked via Gumbel-max sampling with $\tau=0.8$, and the top 6, next 6 and next 8 frames are resized to three scales and fed in temporal order as 20 separate images after a JPEG round trip.
All methods receive the same 32 decoded frames, prompts and scoring protocol; Q-Frame releases no visual tokens.
We seed the Gumbel noise per request for determinism.

\noindent\textbf{Configurations.}
We test three Q-Frame variants on Qwen3-VL and report the most accurate one in Tab.~2.
The official configuration scales retained frames to 1/2, 1/4 and 1/8 of the source dimensions; its token count thus depends on source resolution and becomes very small on TC and MV.
An fp32 ranking, which avoids underflow in the official fp16 softmax, gives nearly the same accuracy.
The resolution-matched variant scales frames to 1, 1/2 and 1/4 of the full model's target input size, spending roughly half the full token count on every benchmark.
This variant is the most accurate (\cref{tab:s-qframe}), and we apply it unchanged to Qwen2.5-VL.

\noindent\textbf{Cost of selection.}
Q-Frame runs slower than the full model on both backbones.
On VM with Qwen3-VL, out of 0.86\,s per request, Long-CLIP CPU preprocessing and scoring take 0.40\,s, resizing, JPEG re-encoding and processor formatting take 0.23\,s, and model inference takes 0.24\,s.
Excluding selection, resizing and the JPEG round trip from the timer raises Q-Frame to 1.48$\times$ TTFT and 1.09$\times$ E2E speedups, which remains slower than ESSENCE-80 at 1.62$\times$ and leaves the matched-speed pairing unchanged.

\noindent\textbf{128 candidates.}
The Q-Frame paper also evaluates selection from 128 candidates, keeping 4, 8 and 32 frames across the three scales.
On VM and LVB with Qwen3-VL, using 128 candidate frames per video (640\,px long side) with resolution matching yields 61.07\% and 59.16\% accuracy at 1598 visual tokens---1.04 ($p=0.22$) and 0.37 points ($p=0.78$) below the 32-candidate setting---while selection alone takes 1.01\,s per request.
ESSENCE-80 outperforms this 128-candidate variant by 3.67 ($p<10^{-4}$) and 2.39 points ($p=0.051$).

\noindent\textbf{Accuracy at matched speed.}
The largest ESSENCE budget at least as fast as Q-Frame is ESSENCE-80 on both models.
On Qwen3-VL, ESSENCE-80 leads Q-Frame by 8.80, 2.63, 6.63 and 2.02 points on TC, VM, MV and LVB ($p<10^{-5}$, $0.0005$, $<10^{-5}$ and $0.09$).
On Qwen2.5-VL, it leads by 3.48, 3.04 and 3.55 points on TC, VM and MV ($p=0.004$, $0.0001$ and $<10^{-5}$) and trails by 0.60 points on LVB ($p=0.64$), where Q-Frame also exceeds the full model by 1.72 points ($p=0.15$).
On average across the four benchmarks, ESSENCE-80 leads by 5.0 points on Qwen3-VL and 2.4 points on Qwen2.5-VL.
At a closer token budget on Qwen3-VL, ESSENCE-66 (66\% tokens vs.\ $\sim$52\% for Q-Frame) gains 8.80, 2.37, 5.87 and 1.20 points ($p<10^{-5}$, 0.002, $<10^{-5}$ and 0.33), and ESSENCE-42 changes accuracy by 8.42, 0.15, 4.76 and $-0.82$ points.
While question-guided selection aids long videos on LVB, Q-Frame loses heavily on TC and MV, which test fine temporal order and state changes, because it feeds 20 selected frames as independent images without timestamps.

\noindent\textbf{LDDR.}
LDDR~\cite{lddr2026} also selects frames before encoding and assigns each retained frame an individual resolution.
We port the official Qwen2.5-VL implementation (commit 33f96f1) to Qwen3-VL and Qwen2.5-VL, keeping its selection and sizing functions intact.
Long-CLIP features of the 32 candidates are weighted by question relevance, a greedy linear DPP selects up to $B/512$ frames, and each selected frame is allocated 256 to 1024 patches proportional to its leave-one-out log-determinant importance and local density.
Selected frames are resized to these patch counts (assuming 14\,px patches) and passed in temporal order as separate images without timestamps or token release.
We match $B$ to the ESSENCE-66 token budget and use Long-CLIP-L at 336\,px as in Q-Frame rather than the default 224\,px.
Because Qwen3-VL uses 16\,px patches, the same pixel dimensions yield about 23\% fewer visual tokens.
LDDR retains about 21 frames per request, using 60--76\% of the full model's visual tokens on Qwen2.5-VL and 63--77\% on Qwen3-VL (\cref{tab:s-qframe}).

\noindent\textbf{Cost of LDDR.}
LDDR is also slower than the full model, achieving 0.52$\times$ and 0.61$\times$ TTFT speedups on Qwen3-VL and Qwen2.5-VL.
During evaluation, DPP selection and CPU preprocessing for Long-CLIP take 0.22 to 0.50\,s per request, and resizing plus processor formatting add 0.09 to 0.19\,s---exceeding the full model's entire TTFT.

\noindent\textbf{LDDR at matched speed.}
The largest budget at least as fast as LDDR is again ESSENCE-80, which runs 3.1 and 2.6 times as fast as LDDR.
On Qwen3-VL, ESSENCE-80 gains 5.13, 0.52, 4.87 and 0.82 points over LDDR on TC, VM, MV and LVB ($p<10^{-5}$, $0.47$, $0.0002$ and $0.46$).
On Qwen2.5-VL, it gains 1.65, 1.74, 1.79 and 0.90 points ($p=0.17$, $0.031$, $0.22$ and $0.45$).
At a comparable token count, ESSENCE-66 gains 5.13, 0.26, 4.11 and 0.00 points on Qwen3-VL ($p<10^{-5}$, $0.73$, $0.003$ and 1.0), and changes accuracy by 0.51, 0.96, 2.24 and $-0.45$ points on Qwen2.5-VL (none significant).
As with Q-Frame, the largest gains appear on the short-event benchmarks TC and MV on Qwen3-VL, whereas on the long-video benchmarks VM and LVB, ESSENCE-66 and LDDR perform within one point of each other on both models.

\noindent\textbf{LDDR with 128 candidates.}
Given the same 128 candidate frames per video, LDDR reaches 65.48\% on VM and 62.98\% on LVB on Qwen3-VL---improving over 32 candidates by 1.26 ($p=0.11$) and 2.24 points ($p=0.048$) and over the 32-frame full model by 1.44 ($p=0.09$) and 3.07 points ($p=0.014$), at 2005 and 1978 visual tokens.
This places it 0.74 and 1.42 points above ESSENCE-80 ($p=0.39$ and $0.26$) and 1.00 and 2.24 points above ESSENCE-66 ($p=0.23$ and $0.07$), with no difference statistically significant.
Note that the inputs are not matched, as ESSENCE and the full model observe only 32 frames: quadrupling the candidate pool benefits LDDR on long videos, whereas it did not help Q-Frame.
Selection alone takes 0.74 and 0.67\,s per request (vs.\ 0.45 and 0.42\,s for 32 candidates), and the selected frames yield about 7\% more visual tokens.

\subsection{Sampling Interval and Video Duration}
\label{sec:s-interval}

\begin{table}[t]
\centering
\scriptsize
\setlength{\tabcolsep}{4pt}
\caption{Accuracy against the full model by sampling interval, in points, at the 66\% budget (56\% on InternVL3.5).}
\label{tab:s-interval}
\begin{tabular}{@{}lrrrr@{}}
\toprule
Method & $\le1$\,s & 1--2\,s & 2--8\,s & $>8$\,s \\
\midrule
\multicolumn{5}{@{}l}{\textit{Qwen3-VL-8B}} \\
ESSENCE-66 & $-$0.72$^{*}$ & +1.21 & +1.98 & +0.22 \\
Uniform downscaling & $-$0.72$^{*}$ & $-$0.69 & $-$1.06 & $-$1.50$^{*}$ \\
FastVID & $-$3.35$^{*}$ & $-$4.14$^{*}$ & $-$4.75$^{*}$ & $-$4.01$^{*}$ \\
FlashVID & $-$2.22$^{*}$ & $-$0.86 & $-$3.03$^{*}$ & $-$2.74$^{*}$ \\
iLLaVA & $-$1.61$^{*}$ & $-$2.41$^{*}$ & $-$0.92 & $-$2.29$^{*}$ \\
EarlyTom & $-$2.77$^{*}$ & $-$3.28$^{*}$ & $-$3.30$^{*}$ & $-$3.19$^{*}$ \\
ShortV & $-$1.46$^{*}$ & $-$1.03 & $-$0.79 & $-$2.17$^{*}$ \\
Mean of five baselines & $-$2.28 & $-$2.34 & $-$2.56 & $-$2.88 \\
\midrule
\multicolumn{5}{@{}l}{\textit{Qwen2.5-VL-7B}} \\
ESSENCE-66 & $-$1.41$^{*}$ & +0.17 & +0.66 & +0.34 \\
Uniform downscaling & $-$1.41$^{*}$ & $-$2.24 & $-$0.66 & $-$1.09$^{*}$ \\
FastVID & $-$3.05$^{*}$ & $-$3.97$^{*}$ & $-$5.01$^{*}$ & $-$2.59$^{*}$ \\
FlashVID & $-$2.18$^{*}$ & $-$2.76$^{*}$ & $-$0.92 & $-$1.50$^{*}$ \\
iLLaVA & $-$1.66$^{*}$ & $-$4.14$^{*}$ & $-$1.85 & $-$1.46$^{*}$ \\
EarlyTom & $-$3.03$^{*}$ & $-$3.28$^{*}$ & $-$5.28$^{*}$ & $-$1.91$^{*}$ \\
ShortV & $-$4.16$^{*}$ & $-$5.00$^{*}$ & $-$3.56$^{*}$ & $-$2.92$^{*}$ \\
Mean of five baselines & $-$2.82 & $-$3.83 & $-$3.32 & $-$2.08 \\
\midrule
\multicolumn{5}{@{}l}{\textit{InternVL3.5-8B}} \\
ESSENCE-56 & $-$0.76$^{*}$ & $-$1.90 & $-$0.66 & $-$0.67 \\
Uniform downscaling & $-$1.76$^{*}$ & $-$2.41 & $-$2.51$^{*}$ & $-$1.01 \\
FastVID & $-$4.94$^{*}$ & $-$1.55 & $-$4.35$^{*}$ & $-$2.17$^{*}$ \\
FlashVID & $-$5.55$^{*}$ & $-$2.93$^{*}$ & $-$4.62$^{*}$ & $-$1.05 \\
iLLaVA & $-$1.87$^{*}$ & +0.52 & $-$2.77$^{*}$ & $-$1.50$^{*}$ \\
EarlyTom & $-$3.79$^{*}$ & $-$3.79$^{*}$ & $-$6.60$^{*}$ & $-$3.26$^{*}$ \\
ShortV & $-$7.32$^{*}$ & $-$4.31$^{*}$ & $-$9.23$^{*}$ & $-$5.59$^{*}$ \\
Mean of five baselines & $-$4.69 & $-$2.41 & $-$5.51 & $-$2.71 \\
\bottomrule
\end{tabular}
\end{table}

\Cref{tab:s-interval} reports the values plotted in Fig.~7a across four interval bins with 5407, 580, 758 and 2667 questions.
They total 9412 of the 9417 questions: five MV clips have fewer than 32 distinct frames, so repeated frames give a median interval of 0, outside the first bin ($0<\Delta t\le1$\,s); elsewhere these questions count as dense.
On Qwen-VL, ESSENCE and uniform downscaling share identical inputs for $\Delta t\le1$\,s and diverge at wider intervals.
Beyond 1\,s, ESSENCE exceeds the full model in every bin, whereas the five-baseline mean drops 2.1 to 3.8 points below it.
On InternVL3.5, ESSENCE-56 stays closest to the full model in three of the four bins.

\begin{table}[t]
\centering
\scriptsize
\setlength{\tabcolsep}{4pt}
\caption{ESSENCE-66 minus uniform downscaling within one benchmark, in answers, with $p$ where the difference is not zero by construction.}
\label{tab:s-within}
\begin{tabular}{@{}llrrr@{}}
\toprule
Benchmark & $\Delta t$ & Questions & Qwen3-VL & Qwen2.5-VL \\
\midrule
VM & $\le1$\,s & 54 & 0 & 0 \\
 & 1--2\,s & 216 & 0 & $+$6 ($p=0.18$) \\
 & $>2$\,s & 2430 & $+$44 ($p=0.011$) & $+$40 ($p=0.013$) \\
\midrule
LVB & $\le1$\,s & 263 & 0 & 0 \\
 & 1--2\,s & 98 & $+$5 ($p=0.27$) & $+$3 ($p=0.61$) \\
 & $>2$\,s & 976 & $+$20 ($p=0.088$) & $+$11 ($p=0.32$) \\
\bottomrule
\end{tabular}
\end{table}

\noindent\textbf{Within one benchmark.}
Because short and long videos largely come from different benchmarks, \cref{tab:s-within} isolates VM and LVB by sampling interval: within each benchmark, ESSENCE-66 improves over uniform downscaling on videos with $\Delta t>2$\,s, significantly so on VM.

\begin{table}[t]
\centering
\scriptsize
\setlength{\tabcolsep}{4pt}
\caption{Accuracy against the full model on the 3425 sparse questions, in points.}
\label{tab:s-long}
\begin{tabular}{@{}lrrr@{}}
\toprule
Method & Qwen3-VL & Qwen2.5-VL & InternVL3.5 \\
\midrule
FastVID & $-4.18^*$ & $-3.12^*$ & $-2.66^*$ \\
FlashVID & $-2.80^*$ & $-1.37^*$ & $-1.84^*$ \\
iLLaVA & $-1.99^*$ & $-1.55^*$ & $-1.78^*$ \\
EarlyTom & $-3.21^*$ & $-2.66^*$ & $-4.00^*$ \\
ShortV & $-1.87^*$ & $-3.07^*$ & $-6.39^*$ \\
Uniform downscaling & $-1.40^*$ & $-0.99^*$ & $-1.34^*$ \\
ESSENCE-80 & $+1.05$ & $+1.43^*$ & -- \\
ESSENCE-66 (-56) & $+0.61$ & $+0.41$ & $-0.67$ \\
ESSENCE-42 (-32) & $-1.40^*$ & $-0.35$ & $-2.22^*$ \\
\bottomrule
\end{tabular}
\end{table}

\noindent\textbf{Sparse videos.}
Across all 3425 sparse questions (\cref{tab:s-long}), all five baselines and uniform downscaling fall significantly below the full model on all three backbones.
By contrast, ESSENCE-66 and -56 show no significant drop ($p=0.29$, 0.50 and 0.19), and ESSENCE-80 exceeds the full model on both Qwen backbones ($p=0.063$ and 0.016).

\begin{table}[t]
\centering
\scriptsize
\setlength{\tabcolsep}{1.9pt}
\caption{Hour-scale subsets: ESSENCE-66 (-56 on InternVL3.5) minus each method, in points. VM long: 30--60\,min, 900 questions. LVB 3600: 15--60\,min, 564 questions.}
\label{tab:s-hour}
\begin{tabular}{@{}llrrrrrr@{}}
\toprule
Model & Subset & Full & FastVID & FlashVID & iLLaVA & EarlyTom & ShortV \\
\midrule
Qwen3 & VM long & $+1.33$ & $+3.78^*$ & $+2.56^*$ & $+4.44^*$ & $+1.89$ & $+4.89^*$ \\
 & LVB 3600 & $+1.42$ & $+2.30$ & $+2.84$ & $+0.89$ & $+4.43^*$ & $-0.71$ \\
Qwen2.5 & VM long & $+0.33$ & $+2.33$ & $+1.78$ & $+1.11$ & $+1.00$ & $+3.78^*$ \\
 & LVB 3600 & $+1.24$ & $+3.55^*$ & $+2.48$ & $+2.48$ & $+2.48$ & $+3.37$ \\
InternVL & VM long & $+0.33$ & $+1.00$ & $-0.44$ & $+0.78$ & $+0.89$ & $+4.89^*$ \\
 & LVB 3600 & $-3.01^*$ & $-0.35$ & $-2.66$ & $-0.35$ & $+2.13$ & $+1.24$ \\
\bottomrule
\end{tabular}
\end{table}

\noindent\textbf{Hour-scale videos.}
Dedicated hour-long benchmarks (LVBench, MLVU, HourVideo) are outside our evaluation scope, so \cref{tab:s-hour} examines the longest duration splits of VM and LVB.
On both Qwen models, ESSENCE-66 scores above the full model on both subsets (not significant) and beats the five baselines in 19 of 20 comparisons, trailing only ShortV on LVB with Qwen3-VL ($p=0.75$).
On InternVL3.5, ESSENCE-56 is 0.33 points above the full model on VM long but 3.01 points below it on LVB 3600 ($p=0.021$), the sole significant drop against the full model in \cref{tab:s-hour}.
Uniform downscaling to the same budget (a 336 tile) loses a similar 3.19 points on this subset ($p=0.041$), with ESSENCE-56 0.18 points ahead ($p=1.0$).
Broken down by LVB reasoning level, ESSENCE-56 and downscaling lose 2.56 and 1.71 points on the 234 perception questions and 3.33 and 4.24 points on the 330 relation questions, with no single category driving the drop.
Because a fixed-size tokenizer does not blend frames, its full model has no double exposure to remove; on these sparse subsets, both pooling and input downscaling trade spatial detail for a smaller token budget.

\subsection{Number of Input Frames}
\label{sec:s-frames}

\begin{table}[t]
\centering
\scriptsize
\setlength{\tabcolsep}{2.5pt}
\caption{Frame count on sparse videos on Qwen3-VL, in points, with $p$ in parentheses. Ours: ESSENCE-66. Down.: uniform downscaling. Speed: mean time per question of the full model over that of ESSENCE-66.}
\label{tab:s-frames}
\begin{tabular}{@{}rrrrrr@{}}
\toprule
$T$ & $n$ & Ours $-$ down. & Ours $-$ full & Down.\ $-$ full & Speed \\
\midrule
32 & 3406 & $+1.88$ (0.002) & $+0.59$ (0.31) & $-1.29$ (0.006) & -- \\
64 & 2779 & $+2.30$ (0.0005) & $+0.76$ (0.26) & $-1.55$ (0.003) & 1.80$\times$ \\
128 & 2674 & $+1.91$ (0.002) & $+0.52$ (0.42) & $-1.38$ (0.008) & 1.90$\times$ \\
256 & 1976 & $+2.07$ (0.005) & -- & -- & -- \\
\bottomrule
\end{tabular}
\end{table}

\Cref{tab:s-frames} provides the values plotted in Fig.~7c.
We evaluate the full model, uniform downscaling and ESSENCE-66 on VM and LVB with Qwen3-VL as frame count increases.
The token budget scales proportionally with $T$ to keep pixels per grid constant, with release fixed at layer 24.
For $T\ge64$, frames are decoded at 640\,px on the long side, so accuracies are comparable only within each $T$.
At $T=256$, we evaluate the 676 videos whose sampling interval exceeds 4\,s at $T=128$ (1979 questions, 1976 sparse) and omit the full model, which would require roughly 32k visual tokens.
ESSENCE-66 maintains an advantage of about 2 points over downscaling up to 256 frames, and at 64 and 128 frames it stays slightly above the full model at 1.8 to 1.9 times its speed.

\subsection{Latency by Stage}
\label{sec:s-stage}

\begin{table}[t]
\centering
\scriptsize
\setlength{\tabcolsep}{4pt}
\caption{Time to first token by stage, in ms, on the 192 timed requests (one round).}
\label{tab:s-stage}
\begin{tabular}{@{}lrrr@{}}
\toprule
Method & Before the LLM & Prefill & Total \\
\midrule
\multicolumn{4}{@{}l}{\textit{Qwen3-VL-8B}} \\
Full model & 119.4 & 280.7 & 400.1 \\
FastVID & 135.2 & 101.4 & 236.6 \\
FlashVID & 156.1 & 97.9 & 254.1 \\
iLLaVA & 365.4 & 251.8 & 617.3 \\
EarlyTom & 125.2 & 102.1 & 227.4 \\
ShortV & 120.3 & 146.1 & 266.4 \\
VTW only & 120.6 & 196.3 & 316.8 \\
ESSENCE-66 & 85.1 & 139.8 & 224.8 \\
\midrule
\multicolumn{4}{@{}l}{\textit{InternVL3.5-8B}} \\
Full model & 538.8 & 805.9 & 1344.7 \\
FastVID & 553.5 & 174.6 & 728.2 \\
VTW only & 527.4 & 741.7 & 1269.1 \\
ESSENCE-56 & 387.7 & 297.3 & 684.9 \\
\bottomrule
\end{tabular}
\end{table}

For Figs.~1c and 9, we record the time at the entry of the first LLM layer in one round of the 192 timed requests and split TTFT into pre-LLM latency and LLM prefill (\cref{tab:s-stage}).
The recording adds minor overhead, so both figures apply these stage proportions to the two-round TTFT totals in Tab.~2.
We report the stage breakdown for all methods on Qwen3-VL and for four of them on InternVL3.5.
ESSENCE cuts pre-LLM latency by 28--29\% on the two profiled models: Qwen3-VL uses smaller encoder inputs, while InternVL3.5 drops redundant frames on dense videos.
FastVID and FlashVID execute the full vision encoder before running their compression modules.

\subsection{Combination with Existing Accelerators}
\label{sec:s-combine}

\begin{table}[t]
\centering
\scriptsize
\setlength{\tabcolsep}{3pt}
\caption{Release on the baselines on Qwen3-VL, in points. Top: the method with release minus the published method. Bottom: ESSENCE-66 minus the method with release.}
\label{tab:s-rel}
\begin{tabular}{@{}lrrrrr@{}}
\toprule
 & TC & VM & MV & LVB & All \\
\midrule
FlashVID with release & $-0.13$ & $+0.19$ & $+0.18$ & $-0.07$ & $+0.10$ \\
FastVID with release & $+0.19$ & $-0.04$ & $-0.08$ & $-0.30$ & $-0.05$ \\
\midrule
Ours vs.\ FlashVID+release & $+2.41^*$ & $+3.56^*$ & $+1.08^*$ & $+1.94$ & $+2.13^*$ \\
Ours vs.\ FastVID+release & $+4.05^*$ & $+5.59^*$ & $+2.11^*$ & $+3.52^*$ & $+3.63^*$ \\
\bottomrule
\end{tabular}
\end{table}

\noindent\textbf{Release on the baselines.}
In Sec.~5.5, we append ESSENCE's layer-24 release on Qwen3-VL to FlashVID and FastVID across all 9417 questions (\cref{tab:s-rel}), disabling FlashVID's own layer-28 pruning.
Mean latency per question on VM and LVB drops from 0.25 to 0.24\,s for FlashVID and from 0.24 to 0.22\,s for FastVID, matching ESSENCE-66 at 0.22\,s.
Adding release shifts baseline accuracy by at most 0.30 points on any benchmark, and ESSENCE-66 still outperforms both across all four benchmarks.

\noindent\textbf{Still-frame encoding as a plug-in.}
Conversely, feeding ESSENCE's still-frame allocation into FastVID, FlashVID and ShortV on Qwen3-VL increases correct answers on VM by 54 ($p=0.004$), 25 ($p=0.16$) and 41 ($p=0.02$), while changing LVB by $+13$, $+14$ and $-9$ answers (none significant).

\subsection{Extension to GLM-4.1V and LLaVA-OneVision}
\label{sec:s-extension}

Sec.~5.6 applies ESSENCE out of the box to two additional architectures, fixing $\tauc$, the release ratio and the allocation rule prior to evaluation.
All variants use the same 32 frames, prompts and scoring on TC, VM and LVB as the main experiments.
Reported times are mean per-question latencies from decoded frames to the first token, measured during accuracy evaluation rather than with the protocol of \cref{sec:s-timing}, and are comparable within each model.

\begin{table}[t]
\centering
\scriptsize
\setlength{\tabcolsep}{4pt}
\caption{GLM-4.1V-9B: correct answers and mean seconds per question, with the speedup over the full model on all 5617 questions.}
\label{tab:s-extension-glm}
\begin{tabular}{@{}lrrrrrrr@{}}
\toprule
 & \multicolumn{2}{c}{TC (1580)} & \multicolumn{2}{c}{VM (2700)} & \multicolumn{2}{c}{LVB (1337)} & \\
\cmidrule(lr){2-3}\cmidrule(lr){4-5}\cmidrule(lr){6-7}
Method & Correct & s & Correct & s & Correct & s & Speedup \\
\midrule
Full model & 1161 & 0.647 & 1579 & 0.675 & 753 & 0.683 & 1.00$\times$ \\
Downscaling & 1155 & 0.310 & 1558 & 0.320 & 739 & 0.326 & 2.10$\times$ \\
ESSENCE-66 & 1154 & 0.310 & 1672 & 0.306 & 786 & 0.317 & 2.16$\times$ \\
\bottomrule
\end{tabular}
\end{table}

\noindent\textbf{GLM-4.1V-9B.}
This model uses a 3D-patch tokenizer with $u=2$, $\etat=0.008$ and 28 pixels per visual token, and an LLM with $L=40$ layers ($\ell^*=27$).
The full model encodes 16 native pairs capped at $448^2$ pixels without release; uniform downscaling caps grids at $365^2$ with release; and ESSENCE-66 applies our rule at the same budget with release (\cref{tab:s-extension-glm}).
ESSENCE-66 gains 161 correct answers over downscaling on VM and LVB combined ($p\approx10^{-5}$)---114 on VM ($p\approx10^{-5}$) and 47 on LVB ($p=0.0003$)---with $-1$ answer on TC ($p=1$).
Across the three benchmarks, 142 of the gained answers come from sparse videos ($p\approx10^{-5}$) and 18 from dense videos ($p=0.009$).
Compared with the full model, ESSENCE-66 gains 93 answers on VM ($p\approx10^{-5}$) and 33 on LVB ($p=0.015$) while losing 7 on TC ($p=0.43$), whereas downscaling loses 41 overall ($p=0.04$).
On sparse videos, the gain over downscaling (142 answers) exceeds the corresponding gains of ESSENCE-66 on Qwen3-VL and Qwen2.5-VL (64 and 51 answers on the sparse videos of VM and LVB, which hold 3406 of the 3425 sparse questions), following the inverse order of $\etat$ (0.008, 0.035 and 0.042); this aligns with smaller $\etat$ causing stronger frame blending, though three models are too few to establish a formal trend.

\begin{table}[t]
\centering
\scriptsize
\setlength{\tabcolsep}{4pt}
\caption{LLaVA-OneVision-7B: correct answers and mean seconds per question, with the speedup over the full model on all 5617 questions.}
\label{tab:s-extension-ov}
\begin{tabular}{@{}lrrrrrrr@{}}
\toprule
 & \multicolumn{2}{c}{TC (1580)} & \multicolumn{2}{c}{VM (2700)} & \multicolumn{2}{c}{LVB (1337)} & \\
\cmidrule(lr){2-3}\cmidrule(lr){4-5}\cmidrule(lr){6-7}
Method & Correct & s & Correct & s & Correct & s & Speedup \\
\midrule
Full model & 1016 & 0.746 & 1511 & 0.811 & 728 & 0.825 & 1.00$\times$ \\
Pool & 992 & 0.409 & 1447 & 0.469 & 715 & 0.474 & 1.76$\times$ \\
Frames & 1021 & 0.277 & 1447 & 0.322 & 709 & 0.306 & 2.61$\times$ \\
ESSENCE-50 & 1021 & 0.277 & 1450 & 0.454 & 720 & 0.429 & 2.00$\times$ \\
\bottomrule
\end{tabular}
\end{table}

\noindent\textbf{LLaVA-OneVision-7B.}
This model uses a fixed-size tokenizer with $u=1$ and 196 tokens per native frame ($27\times27$ patches pooled to $14\times14$), and an LLM with $L=28$ layers ($\ell^*=19$).
The full model encodes 32 native frames without release (6272 tokens).
We compare two half-budget variants with release: Frames retains 16 native frames (3136 tokens), and Pool encodes all 32 frames and pools each to $10\times10$ after the encoder (3200 tokens).
ESSENCE-50 selects Frames when $\Delta t\le2$\,s and Pool otherwise (\cref{tab:s-extension-ov}).
On dense videos, Frames outperforms Pool by 37 answers (2211 questions, 858 videos, $p=0.021$) at 2.67$\times$ the full model's speed, matching the prediction of Eq.~5 for fixed-size tokenizers.
On sparse videos, Pool leads Frames by only 14 answers ($p=0.55$), compared with 82 answers on InternVL3.5 (\cref{tab:s-coverage}); thus, the sparse branch holds only directionally on LLaVA-OneVision, where running Frames everywhere is faster and within 14 answers of ESSENCE-50.
Overall, ESSENCE-50 trails the full model by 64 answers ($p=0.005$).
We leave to future work whether LLaVA-OneVision's smaller native grid ($27\times27$ vs.\ $32\times32$ on InternVL3.5) explains the reduced benefit of pooling.

\section{Limitations}
\label{sec:s-limits}

\noindent\textbf{Task format.}
Our main evaluation uses multiple-choice benchmarks where the first generated token determines the answer, matching our theoretical bound on release (\cref{sec:s-scope}).
Aside from caption generation---where release preserves judged accuracy but shortens Qwen3-VL outputs---we do not evaluate open-ended QA, temporal grounding or multi-turn dialogue.
For long-form or multi-turn applications, disabling release retains a 1.44 to 1.64$\times$ TTFT speedup.

\noindent\textbf{Moments from timestamps.}
Computing $\neff$ relies on a single coherence time $\tauc$ across all videos.
While content-based similarity did not improve over timestamps on our benchmarks (\cref{sec:s-content}), domains with unusually slow or rapid scene changes, such as static surveillance or fast egocentric footage, may benefit from content-adaptive estimation.

\noindent\textbf{Long videos.}
Our primary experiments sample 32 frames.
On the longest LVB split, ESSENCE-56 on InternVL3.5 drops 3.0 points below the full model, matching uniform downscaling at the same budget (\cref{sec:s-interval}), and dedicated hour-scale benchmarks remain to be tested.
On Qwen3-VL, the accuracy gain from allocation persists up to 256 frames (\cref{sec:s-frames}).

\noindent\textbf{Release depth.}
We apply a fixed relative release depth of $2L/3$ across all backbones.
A stricter null-calibrated criterion places the threshold deeper on Qwen2.5-VL and InternVL3.5 (\cref{sec:s-horizon}), trading a smaller speedup for fewer answer flips.

\noindent\textbf{Baselines and hardware.}
Twelve of the 15 baseline--backbone pairs rely on our ports.
Although verified for structure, token budget and determinism, they lack author-specific hyperparameter tuning on each backbone.
Hyperparameters can be sensitive: ShortV drops 6.1 points below the full model on InternVL3.5 at the default 60\% freeze ratio, whereas freezing 30\% restores full-model accuracy at a 1.19$\times$ TTFT speedup (\cref{tab:s-shortv}).
Similarly, given 128 candidate frames (four times ESSENCE's input), LDDR scores 0.7 and 1.4 points above ESSENCE-80 on VM and LVB with Qwen3-VL, though not significantly (\cref{sec:s-qframe}).
Finally, all latencies are measured on a single GPU architecture (RTX 5090) with FlashAttention-2.

\end{document}